\documentclass[preprint,12pt,authoryear]{elsarticle}
\usepackage{times}  
\usepackage{helvet}  
\usepackage{courier}  
\usepackage[hyphens]{url}  
\usepackage{graphicx} 
\usepackage{natbib}  
\usepackage{caption} 
\usepackage{amsmath,amsfonts}
\usepackage{amssymb}
\usepackage{comment}
\usepackage{float} 
\usepackage{subfig} 
\usepackage{balance}
\usepackage{multirow}
\usepackage{footnote}
\usepackage{setspace}
\usepackage{url}
\usepackage{booktabs}
\usepackage{threeparttable}
\usepackage[ruled]{algorithm2e}
\usepackage{algorithmic}
\usepackage[table,xcdraw,dvipsnames]{xcolor}
\usepackage[normalem]{ulem}
\useunder{\uline}{\ul}{}
\DeclareMathOperator{\tr}{tr}
\newtheorem{corollary}{Corollary}

\newtheorem{definition}{Definition}
\usepackage[title]{appendix}
\renewcommand{\appendixpagename}{Appendix}
\usepackage{amssymb}
\usepackage{listings}
\definecolor{mygreen}{rgb}{0,0.6,0}
\definecolor{mygray}{rgb}{0.5,0.5,0.5}
\definecolor{mymauve}{rgb}{0.58,0,0.82}
\definecolor{background}{rgb}{0.95,0.95,0.92}

\DeclareUnicodeCharacter{2002}{\textendash}
\begin{document}
%

\title{CI-GNN: A Granger Causality-Inspired Graph Neural Network for Interpretable Brain Network-Based Psychiatric Diagnosis}

\author{
Kaizhong Zheng$^{1}$, Shujian Yu$^{2,3*}$, Badong Chen$^{1*}$\\
$^{1}$National Key Laboratory of Human-Machine Hybrid Augmented Intelligence, National Engineering Research Center for Visual Information and Applications, and Institute of
Artificial Intelligence and Robotics, Xi’an Jiaotong University, Xi’an, China \\
$^{2}$Department of Computer Science, Vrije Universiteit Amsterdam, Amsterdam, Netherlands\\
$^{3}$Machine Learning Group, UiT - Arctic University of Norway, Troms{\o}, Norway\\
kzzheng@stu.xjtu.edu.cn, yusj9011@gmail.com, chenbd@mail.xjtu.edu.cn\\
}
\begin{abstract}
\begin{quote}
There is a recent trend to leverage the power of graph neural networks (GNNs) for brain-network based psychiatric diagnosis, which, in turn, also motivates an urgent need for psychiatrists to fully understand the decision behavior of the used GNNs. However, most of the existing GNN explainers are either \emph{post-hoc} in which another interpretive model needs to be created to explain a well-trained GNN, or do not consider the causal relationship between the extracted explanation and the decision, such that the explanation itself contains spurious correlations and suffers from weak faithfulness. 
In this work, we propose a granger causality-inspired graph neural network (CI-GNN), a \emph{built-in} interpretable model that is able to identify the most influential subgraph (i.e., functional connectivity within brain regions) that is causally related to the decision (e.g., major depressive disorder patients or healthy controls), without the training of an auxillary interpretive network. 
CI-GNN learns disentangled subgraph-level representations $\alpha$ and $\beta$ that encode, respectively, the causal and non-causal aspects of original graph under a graph variational autoencoder framework, regularized by a conditional mutual information (CMI) constraint. We theoretically justify the validity of the CMI regulation in capturing the causal relationship. We also empirically evaluate the performance of CI-GNN against three baseline GNNs and four state-of-the-art GNN explainers on synthetic data and three large-scale brain disease datasets. We observe that CI-GNN achieves the best performance in a wide range of metrics and provides more reliable and concise explanations which have clinical evidence. The source code and implementation details of CI-GNN are freely available at GitHub repository (https://github.com/ZKZ-Brain/CI-GNN/).



\end{quote}
\end{abstract}


\maketitle

\section{Introduction}

Psychiatric disorders have constituted an extensive social and economic burden for health care systems worldwide~\cite{wittchen2011size}, but the underlying pathological and neural mechanism of the psychiatric disorders still remains uncertain. There are no unified or neuropathological structural traits for psychiatric diagnosis due to the clinical heterogeneity~\cite{goodkind2015identification, lanillos2020review}. Current diagnosis for psychiatric disorders are mainly based on subjective symptoms and signs~\cite{zhang2021identification}, such as insomnia and anxiety, \emph{etc}. However, this way for diagnosis has huge limitations in heavily relying on related symptoms and observational status, which could lead to misdiagnosis and delay the early diagnosis and treatment~\cite{huang2020identifying}.

\begin{figure}[ht!]
  \centering
  \subfloat[Traditional Psychiatric Diagnostic Model]{\includegraphics[scale=0.31]{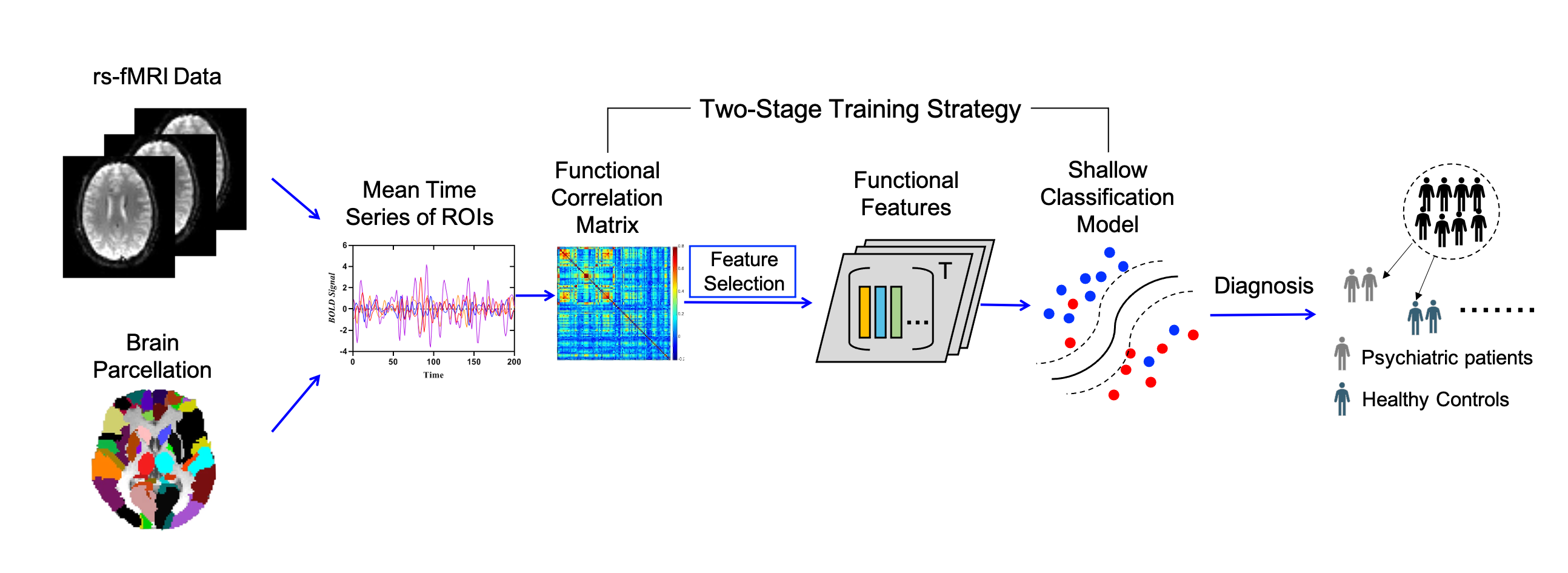}\label{fig:pipeline1}}
  \hfil
  \subfloat[Modern Diagnostic model with \emph{built-in} Interpretable Graph Neural Networks]{\includegraphics[scale=0.34]{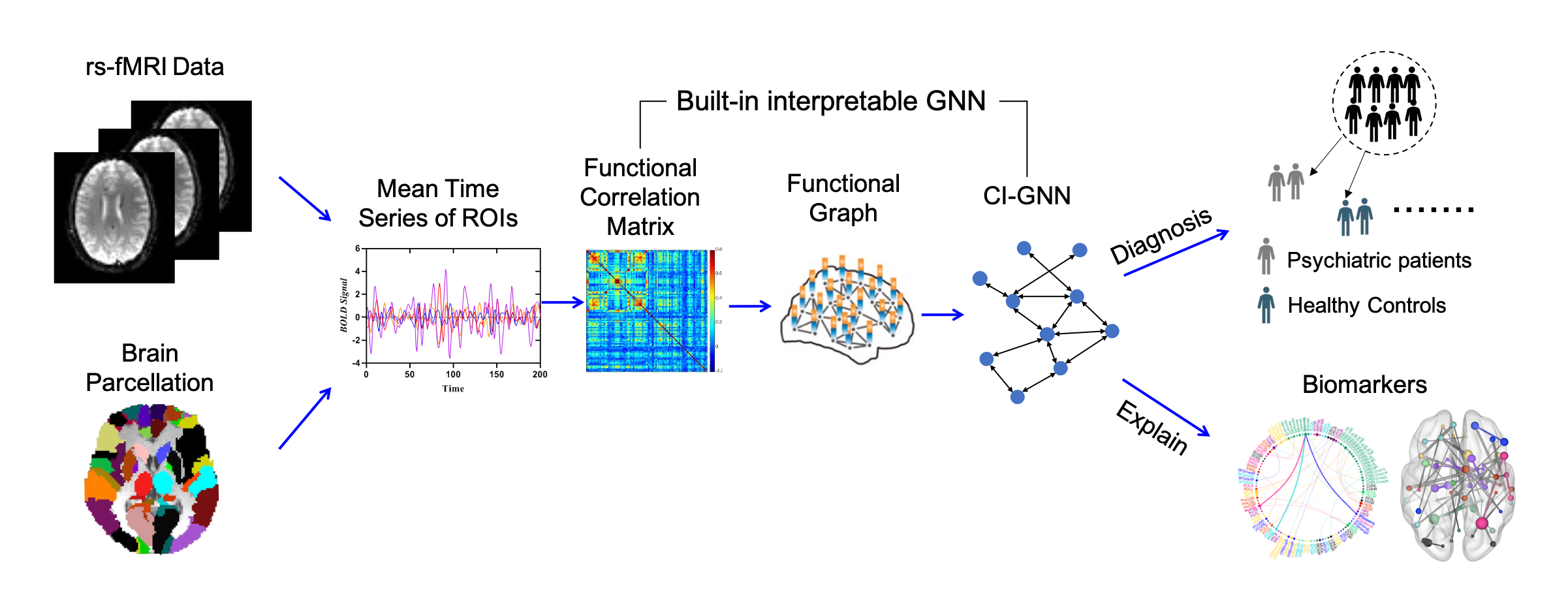}\label{fig:pipeline2}}
  \hfil
  \caption{The overview of the pipeline for (a) traditional psychiatric diagnostic model and (b) modern diagnostic model with built-in interpretable graph neural networks (e.g., our CI-GNN). The resting-state fMRI data are parcellated by an brain atlas such as the automated anatomical labelling (AAL) atlas and calculated the functional connectivity matrices. For traditional psychiatric diagnostic model, which constitutes a two-stage training strategy, it firstly selects the most informative features using feature selection techniques and then discriminates psychiatric patients and healthy controls using classic classification models on top of selected features. For CI-GNN, the functional connectivity matrices are transferred to functional graphs, which are then sent to CI-GNN to make a decision (i.e., psychiatric patients or healthy controls). Our CI-GNN can also discover the most informative edges, $a.k.a.$, potential biomarker, for each participant.}
  \label{fig:pipeline}
\end{figure}

As a noninvasive neuroimaging technique, the functional magnetic resonance imaging (fMRI)~\cite{matthews2004functional} has become popular to investigate neural patterns of brain function for psychiatric disorders~\cite{peraza2020modeling}. Using fMRI, extensive studies in psychiatric diagnosis have been conducted to apply functional connectivity (FC) measured with the pairwise correlations of fMRI time series as features to discriminate psychiatric patients and healthy controls, as illustrated in Figure~\ref{fig:pipeline1}. In general, current diagnostic models based on FC usually adopt a two-stage training strategy which includes feature selection (e.g., two sample \emph{t}-test~\cite{du2018classification}, principal component analysis (PCA)~\cite{zhang2019tensor} and  clustering coefficient~\cite{challis2015gaussian}) and shallow classification model (e.g., support vector machines (SVM)~\cite{pan2018novel}, LASSO~\cite{yamashita2020generalizable} and random forest~\cite{cordova2020heterogeneity}). However, shallow or simple classification models could not capture and analyze the topological and nonlinear information of complex brain networks and fail to achieve acceptable performance for large-scale datasets~\cite{sui2020neuroimaging}. In addition, the performance of these models depend heavily on the used feature selection methods and the classifier, which may lead to inconsistent performance and unreliable or even inaccurate predictions ~\cite{li2021braingnn}. 


Recently, an emerging trend is to utilize graph neural networks (GNNs)~\cite{hamilton2017inductive,zhao2022deep} to construct unified end-to-end psychiatric diagnostic model~\cite{zhou2020toolbox}. In these studies, brain could be modeled as a graph, with nodes representing brain regions of interest (ROIs) and edge representing FC between ROIs.  Despite a substantial improvement in performance, most of existing GNN models are essentially still black-box regimes which are hard to elucidate the underlying decisions behind the predictions. These black-box models cannot be fully trusted and do not meet the demands of fairness, security, and robustness~\cite{yuan2020xgnn,yuan2021explainability,zeng2022shadewatcher}, which severely hinders their real-world applications, particularly in medical diagnosis in which a transparent decision-making process is a prerequisite~\cite{li2021braingnn}. Therefore, developing novel GNNs with both high precision and human understandable explanations is an emerging consensus.

So far, tremendous efforts have been made to improve the explainability of GNNs. According to what types of explanations are provided~\cite{yuan2020explainability}, existing approaches can be divided into two categories: the instance-level explanation and the model-level explanations. For example, GNNExplainer~\cite{ying2019gnnexplainer} and PGExplainer~\cite{luo2020parameterized} extract a compact subgraph to provide the instance-level explanations, while XGNN~\cite{yuan2020xgnn} generates the discriminative graph patterns to provide the model-level explanations. 
Despite of recent advances, existing GNN explainers usually suffer from one or more of the following issues:

\begin{enumerate}
\item{\textbf{\emph{Post-hoc} explanation:} Most explainers are \emph{post-hoc}, in which another interpretive model needs to be created to explain a well-trained GNN~\cite{zhang2022protgnn}. However, \emph{post-hoc} explanations are usually not reliable, inaccurate, and could even be misleading in the entire model decision process~\cite{rudin2018please}. Unlike \emph{post-hoc} explanation, \emph{built-in} interpretable models~\cite{miao2022interpretable} generate explanations by models themselves, without the post-training of an auxiliary network. Hence, the \emph{built-in} explanations are regarded to be more faithful to what the models actually reveal.}

\item{\textbf{Ignorance of causal-effect relationships:} Most GNN explainers recognize predictive subgraphs only by the input-outcome associations rather than their intrinsic causal relationships, which may lead to the obtained explanations contain spurious correlations that are not trustable.}

\item{\textbf{Small-scale evaluations:} In biomedical fields such as bioinformatics and neuroimaging, most GNN explainers are just applied to small-scale datasets, such as molecules~\cite{debnath1991structure} and proteins~\cite{borgwardt2005protein}. In this sense, their practical performance in large-scale biomedical applications (e.g., brain disease diagnosis) is still uncertain.}
\end{enumerate}

To address above technical issues and also evaluate the usefulness of interpretability in real-world clinical applications, we develop a new GNN architecture for psychiatric diagnosis which is able to discriminate psychiatric patients and healthy controls, and obtain biomarkers causally related to the label $Y$, as illustrated in Figure~\ref{fig:pipeline2}.
We term our architecture the Causality Inspired Graph Neural Network (CI-GNN) and evaluate its performance on two large-scale brain disease datasets. Note that, the general idea of explainability of GNNs has recently been extended to brain network analysis.
BrainNNExplainer~\cite{cui2021brainnnexplainer} and IBGNN~\cite{cui2022interpretable} learn a global explanation mask to highlight disorder-specific biomarkers including salient ROIs and important connections, which produces group-level explanations. However, for psychiatric diagnosis, the individual-level explanation is more important, due to the uncertain onset and heterogeneity of symptoms in patients~\cite{qiu2020development}. BrainGNN~\cite{li2021braingnn} designs a novel ROI-aware graph convolutional (Ra-GConv) layers and pooling layer to detect the important nodes for investigating the disorder mechanism, which generates a node-level explanation. However, for brain disorders, it is recognized that connections rather than single nodes alterations can reveal properties of brain disorders~\cite{van2019cross}, which means edge-level (i.e., functional connectivities) explanations are more critical than node-level explanations in brain disorders analysis. Note that, group-level, individual-level, node-level and edge-level are based on differences in the scope of explanation. Therefore, distinct to BrainNNExplainer and BrainGNN, we propose in this work a built-in interpretable GNN for brain disorders analysis that can produce instance-level explanation on edges. Moreover, our explanation is also causal-driven.

To summarize, our main contributions are threefold:
\begin{itemize}
\item{We propose a new built-in interpretable GNN for brain disorders analysis. Our developed CI-GNN enjoys a few unique properties: the ability to produce instance-level explanation on edges; the causal-driven mechanism; and the ability to learn disentangled latent representations $\alpha$ and $\beta$, such that only $\alpha$ induces a subgraph $G_{\text{sub}}$ that is causally related to $Y$.}

\item{We rely on the conditional mutual information (CMI) $I(\alpha;Y|\beta)$ to measure the strength of causal influence of subgraph on labels. We also analyze the rationality of CMI term from a Granger causality perspective~\cite{seth2007granger}. Additionally, we introduce the matrix-based R{\'e}nyi's $\delta $-order mutual information~\cite{giraldo2014measures,yu2019multivariate} to make the computation of the CMI term amenable, such that the whole model can be trained end-to-end.}

\item{Extensive experiments are conducted on synthetic data and two multi-site, large-scale brain disease datasets which contain more than $1,000$ participants across $17$ independent sites, demonstrating the effectiveness and superiority of CI-GNN. In contrast, ~\cite{cui2021brainnnexplainer} only tests on 52 patients with bipolar disorder and 45 healthy controls.
Note that, as demonstrated in the Appendix, our model can also be generalized to other fields such as molecules.}


\end{itemize}

\section{Related work}
\subsection{Psychiatric Diagnostic Model}
The past few years have seen growing prevalence of utilizing FC as neurological biomarkers to develop the computer-aided diagnosic models for psychiatric disorder. Traditionally, these models are two-stage training strategy where some connections are selected from all the connections FC using feature selection methods and concatenated as a long feature vector, and then are sent to classification model. For feature selection methods, one typical approach is to use the group-level statistical test~\cite{du2018classification} such as \emph{t}-test, ranksum-test at each FC edge to select the salient connections that are significantly different between two groups. Another popular approach is to use unsupervised dimension reduction~\cite{zhang2019tensor} such as principal component analysis (PCA), tensor decomposition approach to extract low-dimensional features. \cite{zhang2019tensor} propose a novel tensor network principal components analysis (TN-PCA) method to obtain low-dimensional features from brain network matrices.

For classification model, shallow or linear machine learning classifiers~\cite{rubin2018pattern} such as random forest (RF) and support vector machine (SVM) are still the most popular choices. \cite{rubin2018pattern} use SVM to develop a diagnostic model for bipolar disorder (BD) and major depressive disorder (MDD), and achieves a combined accuracy of 75\%. However, shallow or linear classification models could not capture and analyze the topological and nonlinear information of complex brain networks. More importantly, they fail to achieve acceptable performance for large-scale psychiatric datasets more than 1000 subjects~\cite{sui2020neuroimaging}. In addition, two-stage training strategy could lead to the unreliable and inaccurate predictions. In this study, CI-GNN is a unified end-to-end and selfexplaining disease diagnostic model which is able to identify causal biomarkers elucidating the underlying diagnostic decisions.

\subsection{Graph Neural Networks}
Because graph neural networks (GNNs) have powerful graph-representation ability, they have been widely applied in the various graph tasks including graph classification~\cite{zhang2018end}, link prediction~\cite{schlichtkrull2018modeling} and node classification~\cite{bo2021beyond}. Our brain network analysis deals with graph classification. Let $\left\{ \left(\mathcal{G}_{1}, Y_{1} \right), \left( \mathcal{G}_{2},Y_{2} \right),...,\left(\mathcal{G}_{n},Y_{n}\right)\right\}$ be a set of $n$ graphs with their corresponding labels. Given $\mathcal{G}_{n}=\left(\mathbb{V}, \mathbb{E}\right)$ the $n$-th graph of size $N_{n}$, $\mathbb{V}=\left \{ V_{i} | i\in\left \{1, 2,..., N_{n} \right \}  \right \}$ and $\mathbb{E}=\left \{ \left ( V_{i}, V_{j} \right )| V_{i}, V_{j} \in  \mathbb{V} \right \} $ represent nodes and edges set of $\mathcal{G}_{n}$, respectively, GNN leverages aggregation strategy $\text{A}$ to learn the representation of node $v$ of $\mathcal{G}$ and further use READOUT strategy $\text{R}$ to learn the representation of $\mathcal{G}$.  Their message passing procedures are defined as:

\begin{equation}
h_{v}^{\left ( k \right )}=\text{A}^{\left ( k \right )}\left ( \left\{ h_{u}^{\left ( k-1 \right )} :u \in \mathcal{N} \left ( v \right )\right\} \right ),
\end{equation}

\begin{equation}
h_{G}=\text{R}\left ( \left\{h_{v}^{\left ( k \right )}|v \in G \right\} \right ),
\end{equation}
where $h_{v}^{\left ( k \right )}$ is the representation of node $v$ for the $k$-th layer, $h_{G}$ refers to the representation of entire graph, and  $\mathcal{N} \left ( v \right )$ is the set of neighbour nodes of $v$.

GraphSAGE~\cite{hamilton2017inductive}, Graph Convolutional Network (GCN)~\cite{welling2016semi} and Graph Isomorphism Network (GIN)~\cite{xu2018powerful} are the most popular GNNs, and their aggregation strategy utilize max-, mean- and sum-pooling respectively. For $\text{R}$, a typical implementation is averaging or summation~\cite{welling2016semi,xu2018powerful}. Recently, several studies consider the hierarchical structure of graph (i.e. the pooling aggregation) to learn the graph embedding~\cite{ying2018hierarchical,bianchi2020spectral}.

 In this work, we selected the basic classifier (GCN, GAT, GIN) and READOUT strategy (summation, average, max) based on their performances on different datasets.




\subsection{Interpretability in Graph Neural Networks}
\begin{figure*}[ht!]
\centering
\includegraphics[scale=0.38]{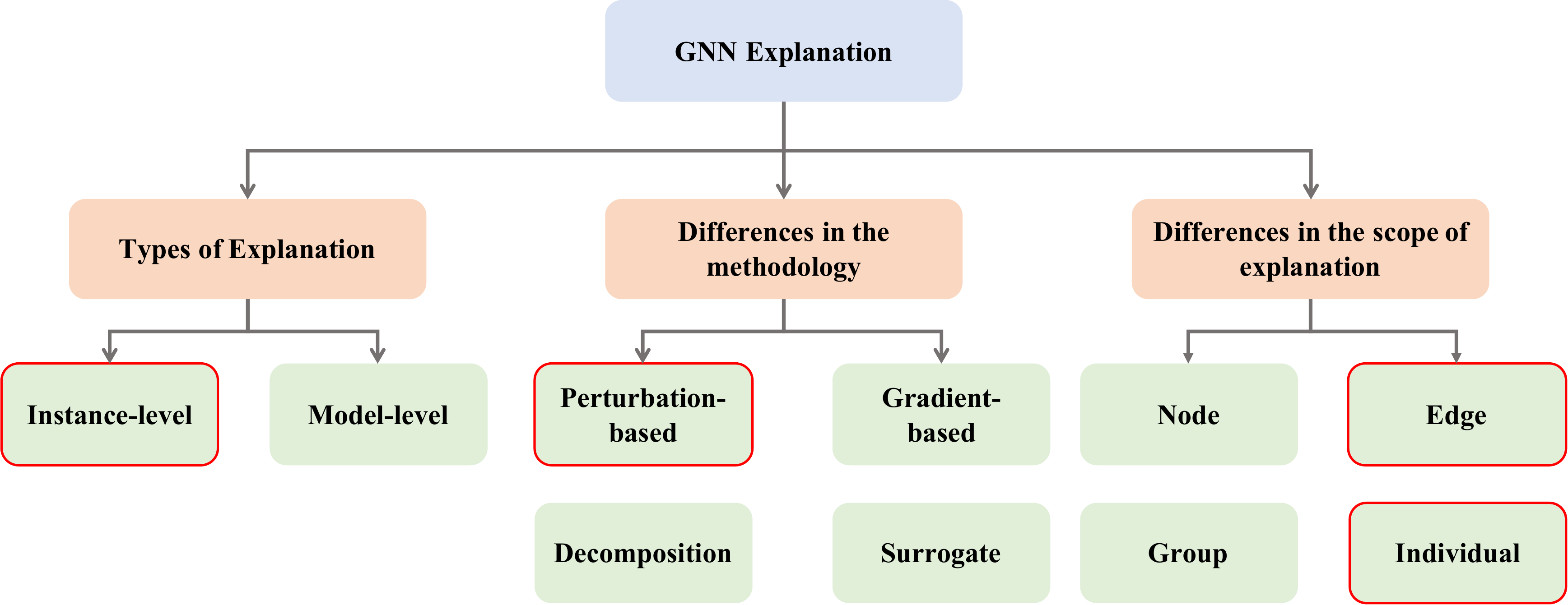}
\caption{ The taxonomy of GNN explanation approaches. The red box is the GNN explanation involved in CI-GNN.
}
\label{fig:explanation}
\end{figure*}

Recently, there is a surge of interest in investigating the GNN explanation. According to differences in the methodology~\cite{yuan2020explainability}, existing GNN explanation approaches can be divided into gradients or features based methods~\cite{baldassarre2019explainability}, perturbation-based methods~\cite{ying2019gnnexplainer}, decomposition methods~\cite{schnake2020xai}, and surrogate methods~\cite{vu2020pgm}. In particular, perturbation-based methods employ distinct mask generators to select important subgraph structure and then evaluate and optimize the mask generators through the performance of subgraphs on a well-trained GNN.  Figure~\ref{fig:explanation} demonstrates the taxonomy of GNN explanation approaches.

The different types of mask generators is an important aspect to distinguish between these methods. For example, GNNExplainer~\cite{ying2019gnnexplainer} and PGExplainer~\cite{luo2020parameterized} employ edge masks to identify a compact subgraph structure, while ZORRO~\cite{funke2020hard} uses node masks and node feature masks to recognize essential input nodes and node features. 

On the other hand, studying the interpretability of GNNs from a causal perspective has recently gained increased attention. StableGNN~\cite{fan2021generalizing} develops the Causal Variable Distinguishing (CVD) regularizer to get rid of spurious
correlation. RC-Explainer~\cite{wang2022reinforced} investigates the causal attribution of different edges to provide local optimal explanations for GNN. Moreover, Gem~\cite{lin2021generative}  leverages the idea of Granger causality to handle the graph-structure data with interdependency, while Orphicx~\cite{lin2022orphicx} further identify the underlying causal factors from latent space.

However, the above mentioned GNN explainers are \emph{post-hoc} in nature and only evaluated on small-scale datasets (such as molecule, proteins and text sentences, etc.) with dozens of nodes. Although DIR-GNN~\cite{wu2021discovering} is also a built-in interpretable GNN based on causal mechanism, it relies on discovering invariant rationales across different distributions, rather than Granger causality.

In this study, we propose a self-explaining GNN model (CI-GNN) which enables automatic identification of subgraphs also from a Granger causality perspective, by learning masks over edges. More importantly, we also exemplify the usefulness of our CI-GNN on two brain disease datasets which contain hundreds of nodes and thousands of edges. Our results as will be shown in the experiments are consistent with clinical observations.

\section{A Granger Causality-Inspired Graph Neural Network}

\begin{figure}[ht!]
  \centering
  \subfloat[]{\includegraphics[scale=0.55]{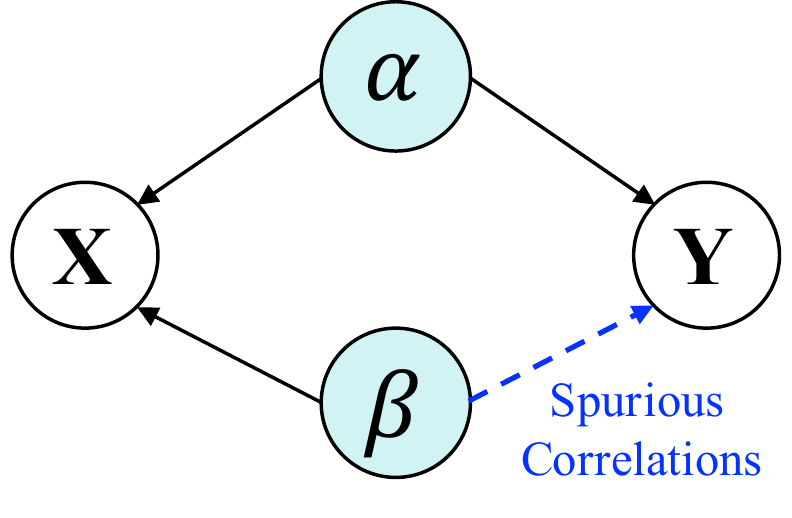}\label{fig:Casual1}}
  \hfil
  \subfloat[]{\includegraphics[scale=0.55]{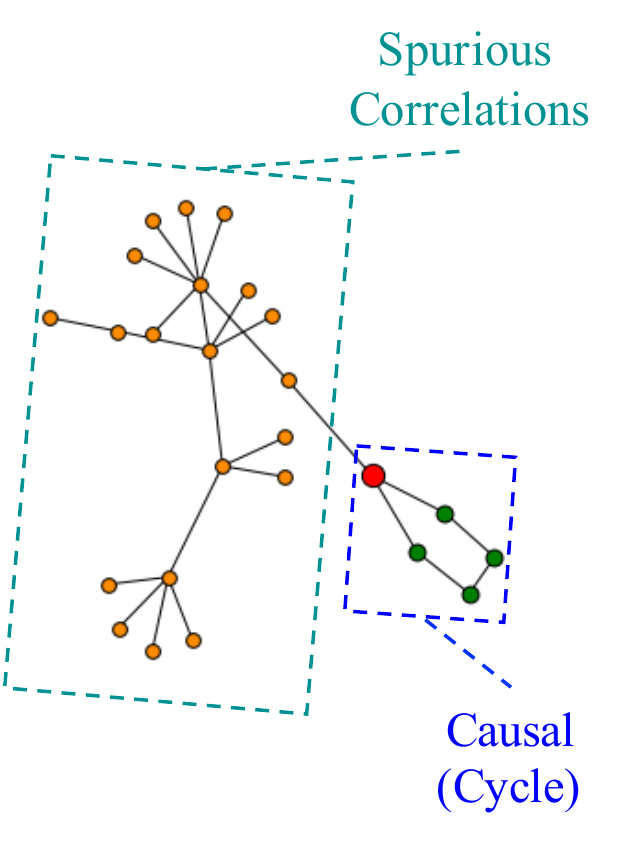}\label{fig:Casual2}}
  \hfil
  \subfloat[]{\includegraphics[scale=0.6]{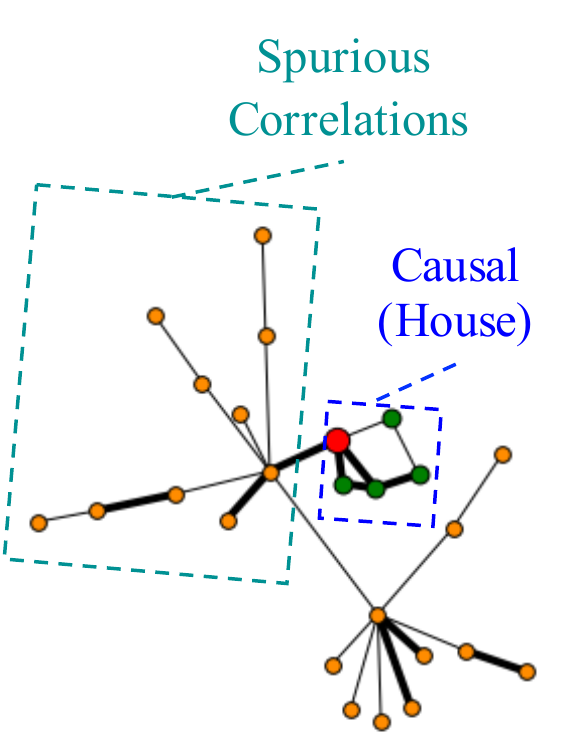}\label{fig:Casual3}}
  \hfil
  \caption{(a) Directed acyclic graph reflects the causal-effect relationship between latent factor $\alpha$ and label $Y$, whereas there is a spurious correlation between $\beta$ and $Y$; (b)-(c) Visualization of causal and non-causal subgraphs for \emph{House} and \emph{Cycle} motif classification. Here, \emph{House} and \emph{Cycle} motifs are causal subgraphs, while \emph{Tree} motif is non-causal subgraph.}
  \label{fig:Casual}
\end{figure}

We use the following causal graph as shown in Figure~\ref{fig:Casual}(a) to elaborate such causal-effect relationship. Specifically, we assume that the input graph data $X$ is generated by latent factors $\alpha$ and $\beta$, whereas only $\alpha$ is causally related to decision or class label $Y$. That is, there is a spurious correlation between $\beta$ and $Y$. 
For example, in the BA-2Motif dataset~\cite{luo2020parameterized} in which graphs with \emph{House} motifs are labeled with $0$ and the ones with \emph{Cycle} are with $1$. Therefore, the remaining subgraph such as the \emph{Tree} structure (see Figure~\ref{fig:Casual}(b-c)) does not ``causally" relate to label $Y$ which can be viewed as spurious correlations.

\begin{figure*}[ht!]
\centering
\includegraphics[scale=0.48]{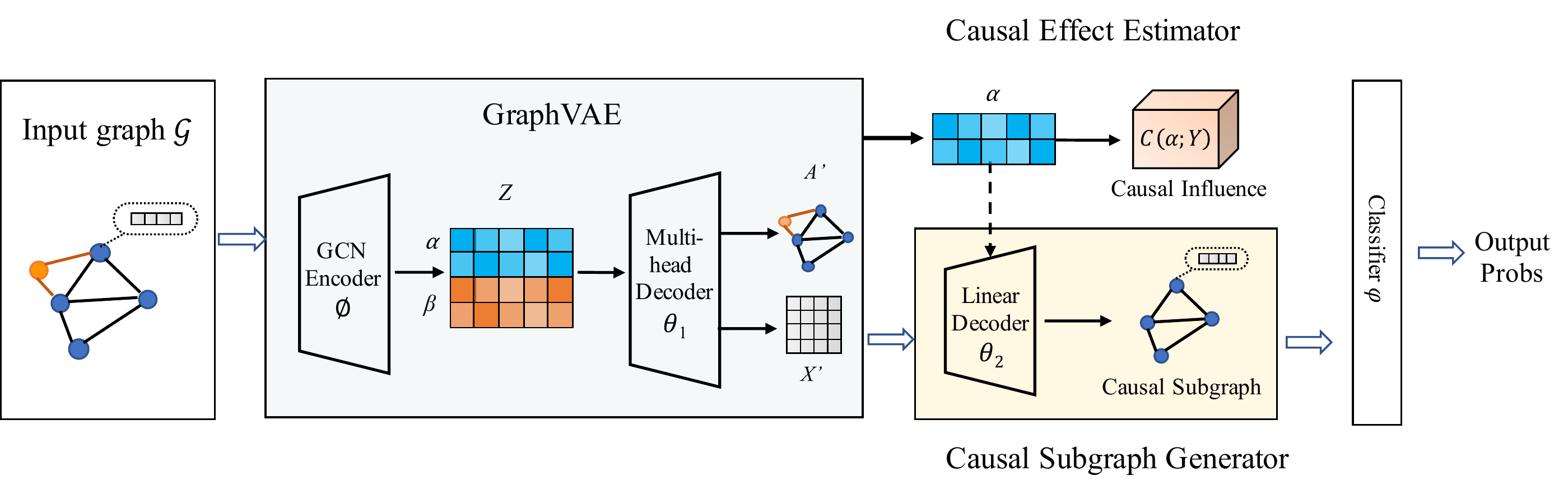}
\caption{ The overall architecture of our proposed CI-GNN. The model consists of four modules: GraphVAE, causal effect estimator, causal subgraph generator and a basic classifier $\varphi$. Given an input Graph $G=\{A,X\}$, GraphVAE learns (disentangled) latent factors $Z=[\alpha;\beta]$. The causal effect estimator ensures that only $\alpha$ is causally related to label $Y$ by the conditional mutual information (CMI) regularization $I\left ( \alpha; Y|\beta \right )$. Based on $\alpha$, we introduce another linear decoder $\theta_2$ to generate causal subgraph $\mathcal{G}_{\text{sub}}$, which can then be used for graph classification by classifier $\varphi$.
}
\label{fig:framework}
\end{figure*}

Based on Figure~\ref{fig:Casual}(a), we develop a causal-inspired Graph Neural Network (CI-GNN\footnote{\url{https://github.com/ZKZ-Brain/CI-GNN/}}). The overall framework of CI-GNN is illustrated in Figure~\ref{fig:framework}. CI-GNN consists of four modules: a Graph variational autoencoder (GraphVAE)~\cite{simonovsky2018graphvae} is used to learn (disentangled) latent factors $Z=[\alpha;\beta]$, from which the decoder is able to reconstruct graph feature matrix $X$ and graph adjancency matrix $A$ with separate heads; a causal effect estimator that ensures only $\alpha$ is causally related to label $Y$; a linear decoder $\theta_2$ to generate the causal subgraph $\mathcal{G}_{\text{sub}}$ from $\alpha$; and a base classifier $\varphi$ that uses $\mathcal{G}_{\text{sub}}$ for graph classification. In the following, we elaborate the four modules in detail.


\subsection{Graph Variational Autoencoders (GraphVAE)}
 Given an input graph $\mathcal{G}=(A,X)$, where $A \in \mathbb{R} ^{n \times n}$ is graph adjacency matrix, node feature $X\in \mathbb{R} ^{n \times d}$ is the node feature matrix, we obtain the reconstructed graph $\mathcal{G}=(A_{c},X_{c})$ using the GraphVAE. Here, $n$ is the number of nodes and $d$ is the dimension of node features. The encoder of GraphVAE is a basic GCN.
Specifically,  $Z$ in the output of a $k$-layer GCN can be defined as:
\begin{equation}
Z^{k} = \sigma  \left ( \tilde{A} Z^{k-1} W^{k-1}\right ),
\end{equation}
where $\tilde{A} = D^{-\frac{1}{2} } \hat{A}   D^{-\frac{1}{2} }$, $\hat{A}=A+I$, $D= {\textstyle \sum_{j}} \hat{A}_{ij}$ is a diagonal degree matrix. In addition, $W$ is a trainable matrix $\sigma\left ( \cdot \right )$ is the sigmoid activation function.


The decoder of our modified GraphVAE has separate heads: a multi-layer perceptron (MLP) is used to reconstruct $X$, and a linear inner product decoder is used to recover $A$. 
Formally, the reconstruction procedure is formulated as:
\begin{equation}
\begin{aligned}
A_{c}=\sigma \left ( ZZ^{T}  \right ) ,\\ 
X_{c}=\text{MLP}\left (Z\right ).
\end{aligned}
\end{equation}

The objective of our GraphVAE can be defined as:
\begin{equation}
\begin{aligned}
 \mathcal{L}_{\text{GraphVAE}}&=\mathbb{E}\left [\left \| X-X_{c} \right \| _{F}\right ]  + \mathbb{E}\left [\left \| A-A_{c} \right \| _{F}\right ] 
 \\&-\mathbb{E} \left [ D_{KL}\left [ q\left ( Z|A,X \right )  \parallel p\left ( Z \right )  \right ] \right ] ,
\end{aligned}
\end{equation}
in which $q\left ( Z|A,X \right )$ denotes the encoder model, $\left \| \cdot \right \| _{F}$ is defined as the Frobenius norm and $p\left ( Z \right )$ represents a prior distribution, which is assumed to follow an isotropic Gaussian.

\subsection{Causal Effect Estimator}
According to Figure~\ref{fig:Casual}(a), $Z\in \mathbb{R} ^{n\times\left (  K+L\right)}$ is consisted of causal factor $\alpha\in \mathbb{R} ^{n\times K}$ and non-causal factor $\beta\in \mathbb{R} ^{n\times L}$. Here, $K$ and $L$ are pre-defined feature dimensions for $\alpha$ and $\beta$, respectively. Our framework needs to ensure that $\alpha$ and $\beta$ are disentangled or independent, and $\alpha$ have a direct causal impact on label $Y$. Mathematically, the objective of the causal effect estimator can be formulated as:
\begin{equation}\label{eq:objective_causal}
\max_{\alpha, \beta} \mathcal{C}\left ( \alpha,Y \right ) -I\left ( \alpha; \beta \right ) ,
\end{equation}
in which $\mathcal{C}\left ( \cdot , \cdot \right )$ measures the strength of causal influence from $\alpha$ to $Y$, $I\left ( \cdot ; \cdot\right )$ denotes mutual information. Minimizing $I(\alpha;\beta)$ forces the independence between $\alpha$ and $\beta$.


According to Corollary~\ref{corollary}, to guarantee $\mathcal{C}\left ( \alpha,Y \right )$ could capture functional dependence, we resort to a conditional mutual information (CMI) term:
\begin{equation}\label{eq:condition}
\mathcal{C}\left ( \alpha,Y \right ) = I\left ( \alpha; Y|\beta \right )
\end{equation}
to measure the causal influence of $\alpha$ on $Y$ when ``imposing" $\beta$. In practice, one can also use the MI term $I(\alpha;Y)$ as suggested in~\cite{o2020generative}. 



\begin{corollary}\label{corollary}
 $I(\alpha;Y|\beta)$ is able to measure the causal effect of $\alpha$ on $Y$ when ``imposing" $\beta$ in the sense of Granger causality~\cite{seth2007granger}. 
\end{corollary}

The proof is shown in the Appendix. Note that, although we do not use ``do” operator to introduce intervention, \cite{o2020generative} shows that $I\left ( \alpha \to Y|do\left ( \beta  \right )  \right )  = I\left ( \alpha; Y|\beta \right )$ from the rules of do-calculus.

Therefore, the objective in Eq.~(\ref{eq:objective_causal}) becomes:
\begin{equation}
\begin{aligned}
\mathcal{L}_{causal} = -I\left ( \alpha; Y|\beta \right ) +I\left ( \alpha; \beta \right ).
\end{aligned}
\end{equation}

\begin{figure}[ht]
\centering
\includegraphics[scale=0.7]{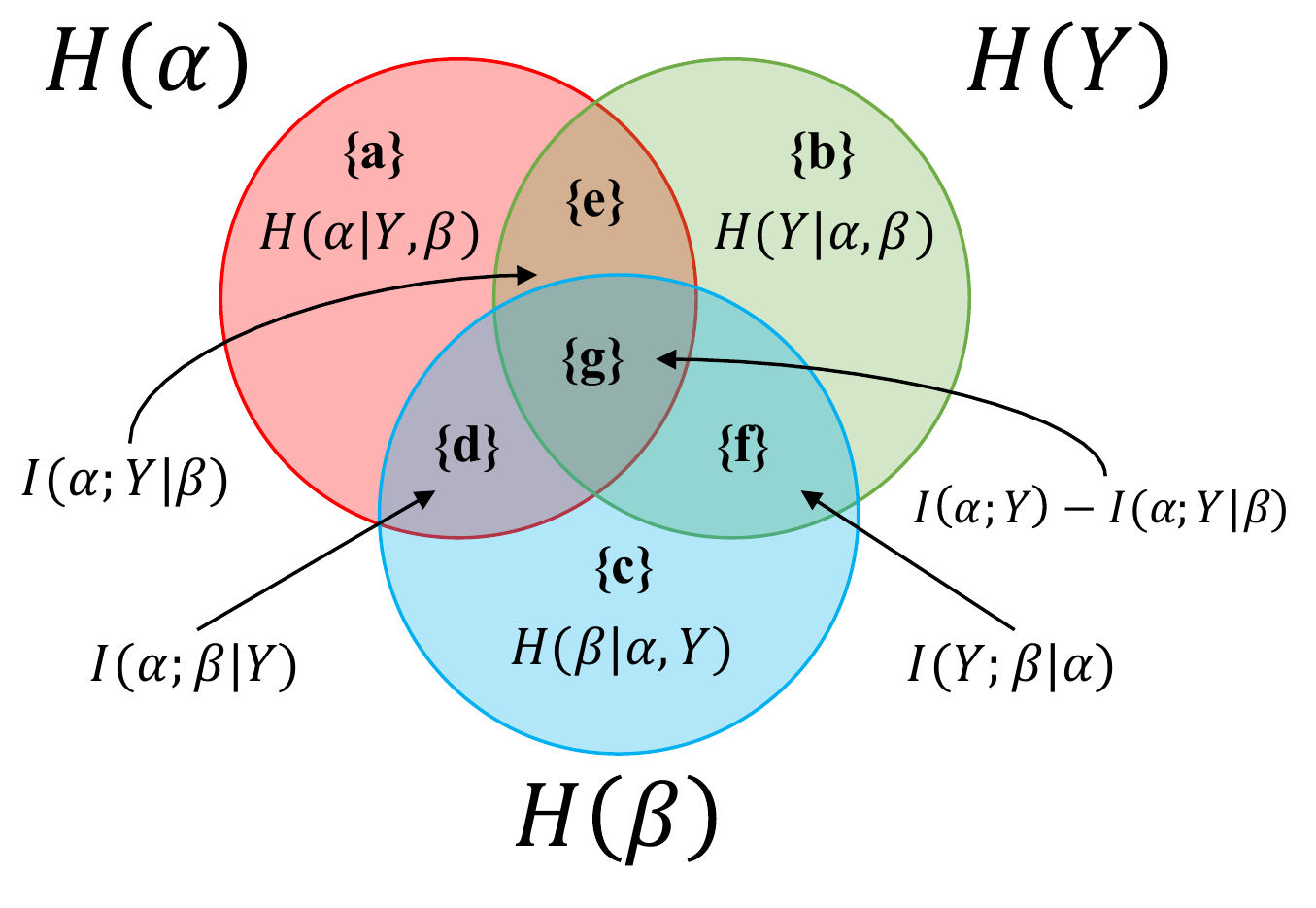}
\caption{Venn diagram depicting entropy interaction among $\alpha$, $\beta$ and $Y$. $H\left ( \alpha \right )=\left \{ a,e,g,d \right \} $, $H\left ( Y \right )=\left \{ b,e,g,f \right \} $, $H\left ( \beta \right )=\left \{ c,d,g,f \right \} $, $I\left ( \alpha; Y \right )=\left \{ e,g \right \} $, $I\left ( \alpha;\beta \right )=\left \{ d,g \right \} $, $I\left ( Y;\beta \right )=\left \{ g,f \right \} $ and $H\left ( \alpha,Y,\beta \right )=\left \{ a,b,c,d,e,f,g \right \} $ }
\label{fig:venn}
\end{figure}

According to Shannon's chain rule~\cite{mackay2003information}, 
$I\left ( \alpha; \beta \right )$ and $I\left ( \alpha; Y|\beta \right )$ can be decomposed as (see also Fig.~\ref{fig:venn} for an illustration):
\begin{equation}\label{eq:MI_decom}
I\left ( \alpha;\beta \right ) =H\left ( \alpha \right ) + H \left ( \beta \right )-H \left ( \alpha,\beta \right ),
\end{equation}
\begin{equation}\label{eq:CMI_decom}
\begin{aligned}
I\left ( \alpha; Y|\beta \right ) = &H\left ( \alpha|\beta \right ) - H\left ( \alpha|Y, \beta \right ) \\= &H\left ( \alpha,\beta \right ) + H\left ( Y,\beta \right )\\
&-H\left ( \beta \right )-H\left ( \alpha,Y,\beta \right ),
\end{aligned}
\end{equation}
in which $H$ denotes entropy or joint entropy. 

In this work, instead of using the popular mutual information neural estimator (MINE)~\cite{belghazi2018mutual} which may make the joint training becomes unstable or even result in negative mutual information values~\cite{yu2022improving}, we
use the matrix-based R{\'e}nyi's $\delta$-order entropy functional~\cite{giraldo2014measures,yu2019multivariate} to estimate different entropy terms in Eqs.~(\ref{eq:MI_decom}) and (\ref{eq:CMI_decom})\footnote{Conventionally, we should use $\alpha$-order. Since $\alpha$ denotes latent causal factor in our work, we use $\delta$-order to avoid confusion.}. This newly proposed estimator can be simply computed (without density estimation or any auxiliary neural network) and is also differentiable which suits well for deep learning applications. For brevity, we directly give the definitions. 


\begin{definition}\label{def_entropy}
Let $\kappa:\chi \times \chi \mapsto \mathbb{R}$ be a real valued positive definite kernel that is also infinitely divisible~\cite{bhatia2006infinitely}. Given $\{\mathbf{x}_{i}\}_{i=1}^{n}\in \chi$, each $\mathbf{x}_i$ can be a real-valued scalar or vector, and the Gram matrix $K\in \mathbb{R}^{n\times n}$ computed as $K_{ij}=\kappa(\mathbf{x}_{i}, \mathbf{x}_{j})$, a matrix-based analogue to R{\'e}nyi's $\delta$-entropy can be given by the following functional:
\begin{equation}\label{Renyi_entropy}
\begin{aligned}
H_{\delta}(A)&=\frac{1}{1-\delta}\log_2 \left(\tr (A^{\delta})\right)\\&=\frac{1}{1-\delta}\log_{2}\left(\sum_{i=1}^{n}\lambda _{i}(A)^{\delta}\right),
\end{aligned}
\end{equation}
where $\delta\in (0,1)\cup(1,\infty)$. $A$ is the normalized $K$, i.e., $A=K/\tr (K)$. $\lambda _{i}(A)$ denotes the $i$-th eigenvalue of $A$.
\end{definition}

\begin{definition}\label{def_joint}
Given a set of $n$ samples $\{\mathbf{x}_{i}, \mathbf{y}_{i},\mathbf{z}_i\}_{i=1}^{n}$, each sample contains three measurements $\mathbf{x}\in\chi$, $\mathbf{y}\in\gamma$ and $\mathbf{z}\in\epsilon$ obtained from the same realization. Given positive definite kernels $\kappa_{1}:\chi\times\chi\mapsto\mathbb{R}$, $\kappa_{2}:\gamma\times\gamma\mapsto\mathbb{R}$, and
$\kappa_{3}:\epsilon\times\epsilon\mapsto\mathbb{R}$,
a matrix-based analogue to R{\'e}nyi's $\delta$-order joint-entropy can be defined as:
\begin{equation}\label{Renyi_joint_entropy}
H_{\delta }(A,B,C)=H_{\delta }\left(\frac{A \circ B \circ C}{\tr (A \circ B \circ C)}\right),
\end{equation}
where $A_{ij}=\kappa_{1}(\mathbf{x}_{i}, \mathbf{x}_{j})$ , $B_{ij}=\kappa_{2}(\mathbf{y}_{i}, \mathbf{y}_{j})$,
$C_{ij}=\kappa_{3}(\mathbf{z}_{i}, \mathbf{z}_{j})$, and $A\circ B \circ C$  denotes the Hadamard product between the matrices $A$, $B$ and $C$.
\end{definition}

Now, given $\left \{ \alpha_{i},\beta_{i},y_{i} \right \}_{i=1}^{B}$ in a mini-batch of $B$ samples, we first need to evaluate three Gram matrices $K_\alpha = \kappa(\alpha_{i}, \alpha_{j}) \in \mathbb{R}^{B\times B}$, $K_\beta = \kappa(\beta_{i}, \beta_{j}) \in \mathbb{R}^{B\times B}$, and $K_y = \kappa(y_{i}, y_{j}) \in \mathbb{R}^{B\times B}$ associated with $\alpha$, $\beta$ and $Y$, respectively. Based on Definitions~\ref{def_entropy} and \ref{def_joint}, the entropy and joint entropy terms in Eqs.~(\ref{eq:MI_decom}) and (\ref{eq:CMI_decom}), such as $H(\alpha)$ and $H(\alpha,Y,\beta)$, all can be simply computed over the eigenspectrum of $K_\alpha$, $K_\beta$, $K_y$, or their Hadamard product. For more details, we refer interested readers to Appendix. 

To obtain causal factor $\alpha$ from $X$,  the overall loss becomes:
\begin{equation}\label{eq:objective_alpha}
\begin{aligned}
\mathcal{L}_{1}= \mathcal{L}_{\text{GraphVAE}} + \lambda\mathcal{L}_{\text{causal}},
\end{aligned}
\end{equation}
where $\lambda$ is the hyper-parameter.

\subsection{Causal Subgraph Generator}
Since $\alpha$ is an embedding representation that is unable to directly provide explanations, we propose causal subgraph generator for better explanation and visualization. To be specific, given 
$\alpha\in \mathbb{R} ^{n\times K}$, we use another linear inner product decoder $\theta_{2}$~\cite{li2021unsupervised} to obtain causal subgraph $\mathcal{G}_{sub}$:
\begin{equation}
\mathcal{G}_{sub}=\sigma \left ( \alpha\alpha^{T}  \right ).
\end{equation}

Subsequently, the trained causal subgraph $\mathcal{G}_{sub}$ is fed into the basic classifier to output the prediction $Y'=\varphi\circ\theta_{2}\left ( X,\alpha \right ) $, where $\varphi$ is a classifier. The cross-entropy loss is used to optimize the basic classifier:
\begin{equation}\label{eq:crossentropy_loss}
\mathcal{L}_{ce}=-\sum_{i=1}^{C} Y_{i}\log \left ( {Y'_{i}} \right )
\end{equation}
in which C is the number of classes.

According to Eq.~(\ref{eq:objective_causal}) and Eq.~(\ref{eq:crossentropy_loss}), the optimization objective of $\theta_{2}$ and $\varphi$ is:
\begin{equation}\label{eq:classification}
\mathcal{L}_{2}= \mathcal{L}_{CE} + \lambda\mathcal{L}_{causal}
\end{equation}

\subsection{Training Procedures}
\begin{algorithm}
\caption{Overview of CI-GNN Training}\label{alg:algorithm}
\LinesNumbered
\KwIn{Training graphs $\mathcal{G}_{train} = \left \{ \mathcal{G}_{i} , y_{i}  \right \}   $, Training epoch $E$, generative casual epoch $E_{GC}$ }
\KwOut{Trained model, Causal subgraph $\mathcal{G}_{\text{sub}}$}
Initialize model parameters.

\For{ $t= 1, 2\cdot\cdot\cdot, E$}{
\eIf{ $t<E_{GC}$}{Optimizing objective function Eq.~(\ref{eq:objective_alpha})\; Performing GraphVAE to generate $\alpha$ and $\beta$}{Optimizing objective function Eq.~(\ref{eq:classification})\; Performing causal subgraph generator to generate $\mathcal{G}_{\text{sub}}$}
}

\end{algorithm}

As shown in Algorithm~\ref{alg:algorithm}, the training procedure includes two stages. We first perform GraphVAE to infer causal factor $\alpha$ and non-causal factor $\beta$ by optimizing Eq.~(\ref{eq:objective_alpha}). When the stage I training is converged (in our implementation, the convergence can be determined if the reconstruction error is smaller than a threshold or the number of training epochs exceeds a predetermined value), we then perform causal subgraph generator to generate $\mathcal{G}_{\text{sub}}$ from $\alpha$ and optimize Eq.~(\ref{eq:classification}).

\section{Experiments}

\subsection{Datasets}
We select one benchmark dataset (BA-2Motif~\cite{yuan2020explainability}) and three brain disease datasets (ABIDE, Rest-meta-MDD and SRPBS) in the experiments:
\begin{itemize}
\item{BA-2Motif~\cite{yuan2020explainability}: It is a graph classification benchmark dataset that contains 1000 graphs and 2 types of graph label. Each graph could attach distinct motifs including \emph{House}-structed motif and \emph{Cycle}-structed motif, which are determined by graph labels.}

\item{ABIDE~\cite{di2014autism}: It openly shares more than 1, 000 resting-state fMRI data of autism patients (ASD) and typically developed (TD) participants\footnote{\url{http://fcon_1000.projects.nitrc.org/indi/abide/}}. Following the preprocessing and sample selection, a total of 528 patients with ASD and 571 TD individuals are used in this paper. }

\item{REST-meta-MDD~\cite{yan2019reduced}: It is one of the largest MDD dataset including more than 2000 participants from twenty-five independent research groups\footnote{\url{http://rfmri.org/REST-meta-MDD/}}. In this study, 1604 participants (848 MDDs and 794 HCs) were selected according to the sample selection.}

 \item{SRPBS~\cite{tanaka2021multi}: It openly shares more than 1000 resting-fMRI data of patients with psychiatric disorders and healthy controls from Japan\footnote{\url{https://bicr-resource.atr.jp/srpbsfc/}}. In this study, 184 participants (92 patients with Schizophrenia and 92 HCs) were selected according to the sample selection.}

\end{itemize}

Table~\ref{tab:Stastics} demonstrates the detailed statistics of these datasets. Preprocessing of three brain disease datasets are provided in the Appendix.

\begin{table}[ht!]
\centering
\renewcommand\arraystretch{1}
\caption{Data statistics.}
\begin{threeparttable}
\setlength{\tabcolsep}{1mm}{}{
\begin{tabular}{c|cccc}
\toprule
Datasets      &  Edges & Nodes & Graphs & Classes \\ \midrule
BA-2Motifs    & 25.48            & 25               & 1000        & 2            \\
ABIDE         & 1334          & 116              & 1099        & 2            \\
REST-meta-MDD & 1334          & 116              & 1604        & 2            \\ 
 SRPBS &  1334          & 116             & 184       & 2          \\\bottomrule
\end{tabular}}
\end{threeparttable}
\label{tab:Stastics}
\end{table}

\subsection{Baselines}
For comparison, we evaluate the performance of our CI-GNN against four traditional psychiatric diagnostic classifiers including support vector machines (SVM) with linear and RBF kernel~\cite{pan2018novel}, random forest (RF)~\cite{rigatti2017random} and LASSO~\cite{ranstam2018lasso}, three baseline GNNs (GCN~\cite{welling2016semi}, GAT~\cite{velivckovic2018graph}, GIN~\cite{xu2018powerful}) and six recently proposed state-of-the-art (SOTA) graph explainers, namely subgraph information bottleneck (SIB)~\cite{yu2021recognizing}, GNNExplainer~\cite{ying2019gnnexplainer}, PGExplainer~\cite{luo2020parameterized}, RC-Explainer~\cite{wang2022reinforced}, OrphicX~\cite{lin2022orphicx} and DIR-GNN~\cite{wu2021discovering}. Note that, SIB also uses information bottleneck to extract subgraph, but it is not causality-driven and does not consider latent factors. OrphicX also originates from a Granger causality perspective and learns disentangled representations $\alpha$ and $\beta$, but it is \emph{post-hoc} and does not ensure the independence between $\alpha$ and $\beta$. In addition, although DIR-GNN~\cite{wu2021discovering} is also a built-in interpretable GNN based on causal mechanism, it relies on discovering invariant rationales across different distributions, rather than Granger causality.

\subsubsection{Implementation Details}
For all competing models, we use the Adam optimizer~\cite{kingma2015adam} and the learning rate is turned in $0.001$. The dropout rate is set as $0.5$ and the weight decay is set as $0.0005$. The number of layers of GCN, GAT and GIN is set as $3$. In addition, the batch size is set as $32$. For CI-GNN, $\lambda$ in Eq.~(\ref{eq:objective_alpha}) is set to be $0.001$, respectively. The feature dimensions $K$ and $L$ for $\alpha$ and $\beta$ are set as $56$ and $8$. In the training of CI-GNN, the train epoch $E$ is set to be $450$, and the generative causal epoch $E_{GC}$ is set to be $150$. For the matrix-based R\'enyi's $\delta$-order entropy, we set $\delta$=1.01 and kernel size $\sigma$ with the average of mean values for all samples as performed in~\cite{zheng2022brainib}. Hyper-parameters in CI-GNN are besed on grid search or recommended settings of related work.

For baselines, we conduct grid search or recommended settings to determine the final settings. We train each model with $300$ epochs. For GIN, GAT and GCN, we use the recommended hyperparameters of related work to train the models. For SIB, the weight $\beta$ of the mutual information term $I\left(\mathcal{G},\mathcal{G}_{sub}\right)$ is selected from $\left \{ 0.00001,0.1 \right \}$. For GNNExplainer, the weight of mutual information (MI) is set to be $0.5$ according to recommended setting. For PGExplainer, the  temperature $\tau$ in reparameterization is set to be $0.1$ according to recommended setting. For RC-Explainer, the beam search is selected from $\left \{ 2,4,8,16 \right \}$.

\subsection{Evaluation on Classification Performance}

\begin{table*}[ht!]
\centering
\caption{ The classification performance and standard deviations of CI-GNN and the baselines on BA-2motifs.The best and second best performances are in bold and underlined, respectively.}
\renewcommand\arraystretch{1}
\begin{tabular}{@{}cccc@{}}
\toprule
\textbf{Method}     & \textbf{Accuracy}    & \textbf{F1}          & \textbf{MCC}         \\ \midrule
GCN                 & 0.50 ± 0.02          & 0.50 ± 0.02          & 0.00 ± 0.00          \\
GAT                 & 0.47 ± 0.02          & 0.47 ± 0.02          & 0.00 ± 0.00          \\
GIN                 & {\ul 0.96 ± 0.02}    & {\ul 0.96 ± 0.02}    & {\ul 0.93 ± 0.03}    \\
SIB                 & 0.53 ± 0.01          & 0.53 ± 0.01          & 0.03 ± 0.03          \\
DIR-GNN             & \textbf{0.99 ± 0.01} & \textbf{0.99 ± 0.01} & \textbf{0.99 ± 0.01} \\
GNNExplainer        & 0.61 ± 0.01          & 0.61 ± 0.01          & 0.26 ± 0.02          \\
PGExplainer         & 0.66 ± 0.01          & 0.66 ± 0.01          & 0.31 ± 0.02          \\
RC-Explainer        & 0.82 ± 0.05          & 0.82 ± 0.05          & 0.64 ± 0.10          \\
CI-GNN (Ours) & \textbf{0.99 ± 0.01} & \textbf{0.99 ± 0.01} & \textbf{0.99 ± 0.01} \\ \bottomrule
\end{tabular}
\label{tab:BAPerformance}
\end{table*}

\begin{table*}[ht!]
\centering
\renewcommand\arraystretch{1.2}
\caption{ The classification performance and standard deviations of CI-GNN and the baselines on three brain disease datasets.The best and second best performances are in bold and underlined, respectively.}
\resizebox{\linewidth}{!}{\begin{tabular}{@{}c|ccc|ccc|ccc@{}}
\toprule
\multirow{2}{*}{Methods} & \multicolumn{3}{c|}{ABIDE}                                         & \multicolumn{3}{c|}{Rest-meta-MDD}                                 & \multicolumn{3}{c}{SRPBS}                                          \\ \cmidrule(l){2-10} 
                         & Accuracy             & F1                   & MCC                  & Accuracy             & F1                   & MCC                  & Accuracy             & F1                   & MCC                  \\ \midrule
Linear-SVM               & 0.67 ± 0.05          & 0.67 ± 0.05          & 0.34 ± 0.11          & 0.63 ± 0.02          & 0.61 ± 0.03          & 0.34 ± 0.14          & 0.87 ± 0.02          & 0.88 ± 0.02          & 0.75 ± 0.03          \\
RBF-SVM                  & {\ul 0.69 ± 0.01}    & {\ul 0.69 ± 0.01}    & {\ul 0.39 ± 0.02}    & 0.66 ± 0.03          & 0.64 ± 0.03          & 0.32 ± 0.05          & 0.87 ± 0.02          & {\ul 0.89 ± 0.01}    & 0.76 ± 0.04          \\
RF                       & 0.64 ± 0.01          & 0.64 ± 0.01          & 0.29 ± 0.05          & 0.60 ± 0.02          & 0.56 ± 0.02          & 0.20 ± 0.03          & 0.84 ± 0.10          & 0.84 ± 0.09          & 0.67 ± 0.20          \\
LASSO                    & 0.65 ± 0.03          & 0.64 ± 0.01          & 0.29 ± 0.05          & 0.61 ± 0.02          & 0.59 ± 0.02          & 0.22 ± 0.04          & 0.79 ± 0.06          & 0.79 ± 0.07          & 0.58 ± 0.13          \\
GCN                      & 0.66 ± 0.06          & 0.65 ± 0.08          & 0.30 ± 0.01          & 0.64 ± 0.04          & 0.65 ± 0.05          & 0.28 ± 0.07          & 0.81 ± 0.08          & 0.81 ± 0.10          & 0.62 ± 0.17          \\
GAT                      & 0.68 ± 0.03          & {\ul 0.69 ± 0.04}    & 0.37 ± 0.07          & 0.64 ± 0.03          & 0.64 ± 0.09          & 0.26 ± 0.05          & 0.84 ± 0.11          & 0.84 ± 0.10          & 0.70 ± 0.19          \\
GIN                      & 0.67 ± 0.04          & 0.67 ± 0.04          & 0.36 ± 0.07          & 0.64 ± 0.02          & 0.63 ± 0.07          & 0.27 ± 0.05          & 0.82 ± 0.08          & 0.80 ± 0.09          & 0.65 ± 0.17          \\
SIB                      & 0.65 ± 0.01          & 0.62 ± 0.01          & 0.29 ± 0.02          & 0.65 ± 0.01          & 0.62 ± 0.01          & 0.29 ± 0.02          & 0.70 ± 0.03          & 0.69 ± 0.04          & 0.43 ± 0.07          \\
DIR-GNN                  & 0.68 ± 0.01          & 0.63 ± 0.02          & 0.35 ± 0.01          & {\ul 0.68 ± 0.01}    & 0.63 ± 0.02          & {\ul 0.35 ± 0.01}    & 0.87 ± 0.03          & 0.88 ± 0.05          & 0.76 ± 0.07          \\
GNNExplainer             & 0.64 ± 0.01          & 0.59 ± 0.06          & 0.28 ± 0.02          & 0.64 ± 0.01          & 0.59 ± 0.06          & 0.28 ± 0.02          & 0.88 ± 0.08          & 0.85 ± 0.10          & 0.75 ± 0.17          \\
PGExplainer              & 0.65 ± 0.03          & 0.67 ± 0.01          & 0.31 ± 0.06          & 0.65 ± 0.03          & {\ul 0.67 ± 0.01}    & 0.31 ± 0.06          & 0.88 ± 0.08          & 0.87 ± 0.11          & 0.75 ± 0.17          \\
RC-Explainer             & 0.68 ± 0.05          & 0.62 ± 0.05          & 0.35 ± 0.11          & 0.64 ± 0.02          & 0.60 ± 0.02          & 0.28 ± 0.05          & {\ul 0.89 ± 0.05}    & 0.88 ± 0.06          & {\ul 0.80 ± 0.11}    \\
CI-GNN(Ours)             & \textbf{0.71 ± 0.02} & \textbf{0.72 ± 0.03} & \textbf{0.43 ± 0.04} & \textbf{0.72 ± 0.02} & \textbf{0.70 ± 0.01} & \textbf{0.45 ± 0.03} & \textbf{0.93 ± 0.03} & \textbf{0.93 ± 0.03} & \textbf{0.86 ± 0.06} \\ \bottomrule
\end{tabular}}
\label{tab:Performance}
\end{table*}

{ We demonstrate the classification performances in terms of Accuracy, F1-score and matthew’s cerrelatien ceefficient (MCC) in Table~\ref{tab:BAPerformance}-\ref{tab:Performance}.  The random data splitting strategy is 80\% for training, 10\% for validation, and the remaining 10\% for testing. Based on the random data splitting, each model was conducted across 3 independent runs. The ablation study, statistical tests over different performance metrics are provided in the Appendix.

 Extensive experiments show that CI-GNN yields significant improvements over all baselines in terms of all evaluating metrics in four graph classification datasets, indicating that CI-GNN has great advantages for graph classification tasks. 


For synthetic dataset, CI-GNN achieves an accuracy rate of 99.9\% exceeding other baselines. Considering the \emph{House} and \emph{Cycle} motif are directly related to labels, the superior performance can be explained by the model's ability to correctly identify causal subgraphs (\emph{House} and \emph{Cycle} motif). This result also suggests that CI-GNN is able to remove the impact of spurious correlation to a certain extent and recognizes causal subgraphs associated with the true label.

For brain disease datasets, CI-GNN achieves significant and consistent improvements over all SOTA approaches. One should note that, ABIDE and REST-meta-MDD are two multi-site, large-scale brain disease datasets including more than $1,000$ participants and $17$ independent sites, where each graph contains hundreds of nodes and thousands of edges. We also performed leave-one-site-out cross validations to further justify the generalization ability of CI-GNN. Empirical results are shown in the Appendix.

\subsection{Hyperparameter Analysis}
We investigate the sensitivity of the pre-defined feature dimensions $K$ and $L$ for $\alpha$ and $\beta$ using ABIDE and REST-meta-MDD datasets. We train the CI-GNN with different ratios of $\frac{K}{K+L}$, when we fix the sum of feature dimensions $Z=K+L$ equals to $64$. As shown in Figure~\ref{fig:Hyperparameter}, we observe that CI-GNN achieves the highest accuracy on both brain disease datasets, when $\frac{K}{K+L}$ is set to $0.8$.

\begin{figure}[ht]
  \centering
  \subfloat[ABIDE]{\includegraphics[scale=0.6]{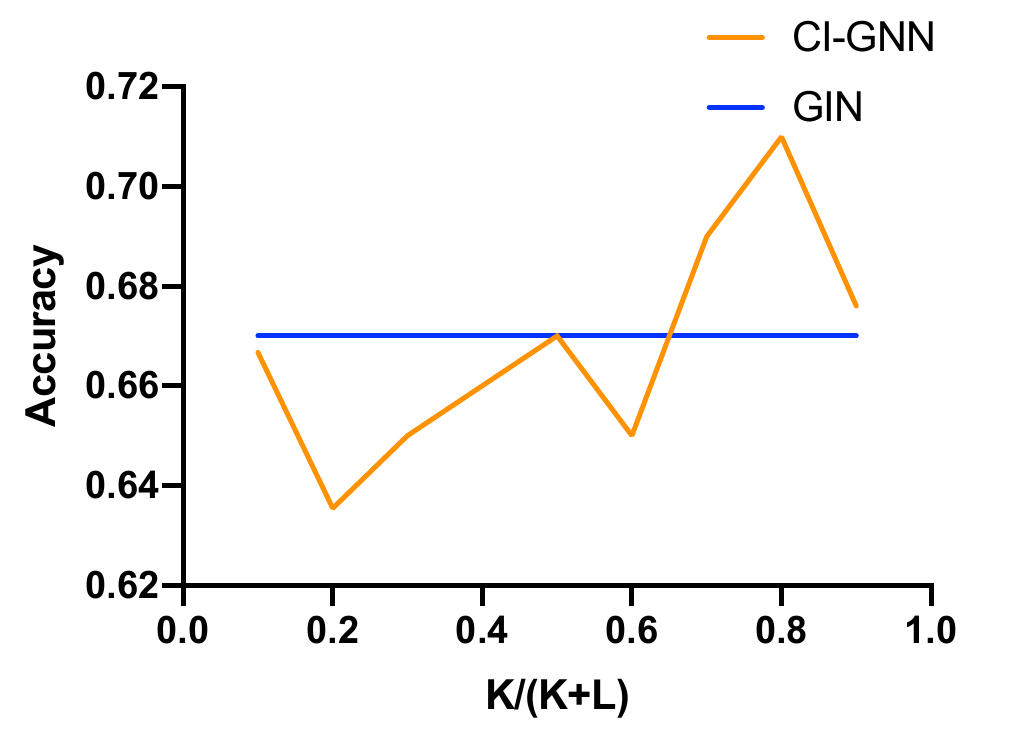}}
  \hfil
  \subfloat[REST-meta-MDD]{\includegraphics[scale=0.6]{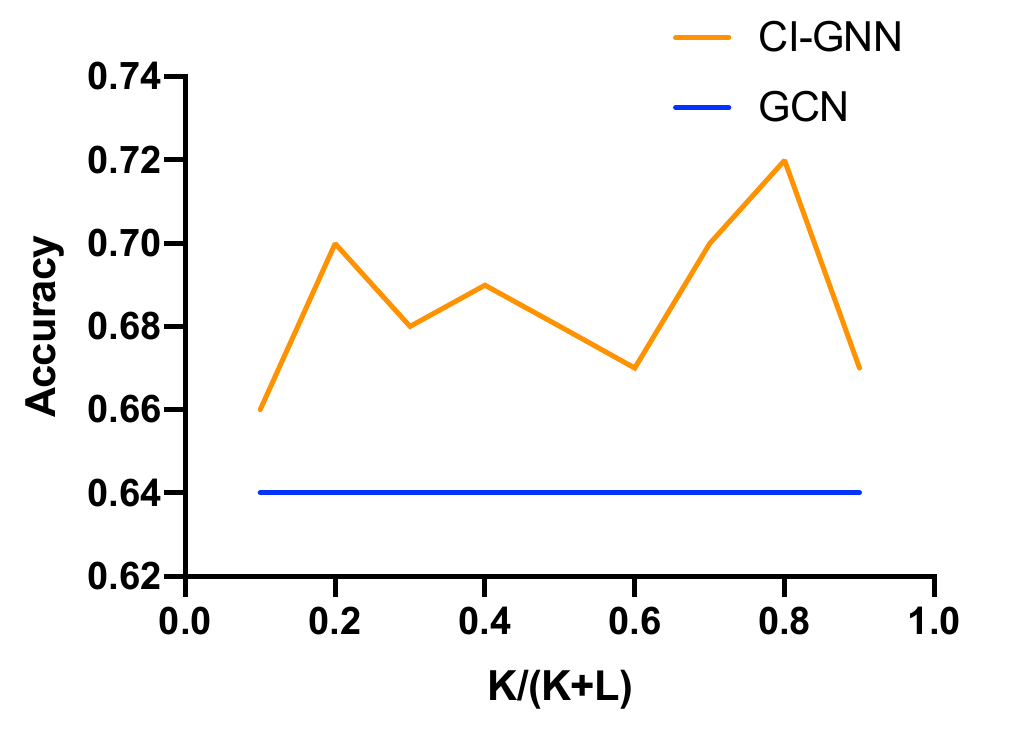}}
  \hfil
  \caption{Sensitivity of the pre-defined feature dimensions $K$ and $L$ for $\alpha$ and $\beta$.}
  \label{fig:Hyperparameter}
\end{figure}

 Furthermore, we further investigate the influence of different basic classifiers and (GCN, GAT, GIN) and READOUT strategies (summation, average, max) on classification performance. Table \ref{fig:Classifier}-\ref{fig:Strategy} show the classification performance of accuracy, F1 and MCC of CI-GNN variants on four datasets. Thus, GIN and summation on BA-2motifs and ABIDE, GCN and summation on Rest-meta-MDD, GIN and average on SRPBS are selected based on the best performance.

\begin{table}[ht!]
\caption{ The classification performance of accuracy, F1 and MCC and standard deviations of CI-GNN variants with different basic classifiers. The best performance is in bold.}
\renewcommand\arraystretch{1}
\resizebox{\linewidth}{!}{\begin{tabular}{@{}c|ccccc@{}}
\toprule
\textbf{Metrics}          & \textbf{Method} & \textbf{BA-2motifs}  & \textbf{ABIDE}       & \textbf{Rest-meta-MDD} & \textbf{SRPBS}       \\ \midrule
\multirow{3}{*}{Accuracy} & CI-GNN (GCN)    & 0.62 ± 0.02          & 0.69 ± 0.02          & \textbf{0.72 ± 0.02}   & 0.84 ± 0.05          \\
                          & CI-GNN (GAT)    & 0.65 ± 0.02          & 0.66 ± 0.02          & 0.66 ± 0.02            & 0.84 ± 0.01          \\
                          & CI-GNN (GIN)    & 0.99 ± 0.01          & \textbf{0.71 ± 0.02} & 0.66 ± 0.03            & 0.86 ± 0.03          \\ \midrule
\multirow{3}{*}{F1}       & CI-GNN (GCN)    & 0.61 ± 0.10          & 0.67 ± 0.08          & 0.70 ± 0.01            & 0.83 ± 0.06          \\
                          & CI-GNN (GAT)    & 0.62 ± 0.01          & 0.63 ± 0.03          & 0.65 ± 0.06            & 0.82 ± 0.01          \\
                          & CI-GNN (GIN)    & \textbf{0.99 ± 0.01} & 0.72 ± 0.03          & 0.67 ± 0.01            & 0.85 ± 0.04          \\ \midrule
\multirow{3}{*}{MCC}      & CI-GNN (GCN)    & 0.25 ± 0.04          & 0.37 ± 0.03          & 0.45 ± 0.03            & 0.69 ± 0.10          \\
                          & CI-GNN (GAT)    & 0.30 ± 0.01          & 0.31 ± 0.05          & 0.31 ± 0.03            & 0.69 ± 0.01          \\
                          & CI-GNN (GIN)    & \textbf{0.99 ± 0.01} & \textbf{0.43 ± 0.04} & 0.32 ± 0.06            & \textbf{0.72 ± 0.06} \\ \bottomrule
\end{tabular}}  
\label{fig:Classifier}
\end{table}

\begin{table}[ht!]
\caption{The classification performance of accuracy, F1 and MCC and standard deviations of CI-GNN variants with different READOUT strategies. The best performance is in bold.}
\renewcommand\arraystretch{1}
\resizebox{\linewidth}{!}{
\begin{tabular}{@{}c|ccccc@{}}
\toprule
\textbf{Metrics}          & \textbf{Method} & \textbf{BA-2motifs}  & \textbf{ABIDE}       & \textbf{Rest-meta-MDD} & \textbf{SRPBS}       \\ \midrule
\multirow{3}{*}{Accuracy} & CI-GNN (Sum)    & 0.99 ± 0.01       & \textbf{0.71 ± 0.02} & \textbf{0.72 ± 0.02}   & 0.86 ± 0.03          \\
                          & CI-GNN (Ave)    & \textbf{0.99 ± 0.01} & 0.66 ± 0.04          & 0.66 ± 0.01            & \textbf{0.93 ± 0.03} \\
                          & CI-GNN (Max)    & \textbf{0.99 ± 0.01} & \textbf{0.66 ± 0.01} & 0.66 ± 0.02            & 0.84 ± 0.01          \\ \midrule
\multirow{3}{*}{F1}       & CI-GNN (Sum)    & \textbf{0.99 ± 0.01} & \textbf{0.72 ± 0.03} & \textbf{0.70 ± 0.01}   & 0.85 ± 0.04          \\
                          & CI-GNN (Ave)    & \textbf{0.99 ± 0.01} & 0.67 ± 0.04          & 0.67 ± 0.02            & 0.93 ± 0.03          \\
                          & CI-GNN (Max)    & \textbf{0.99 ± 0.01} & 0.64 ± 0.01          & 0.65 ± 0.02            & 0.86 ± 0.01          \\ \midrule
\multirow{3}{*}{MCC}      & CI-GNN (Sum)    & \textbf{0.99 ± 0.01} & \textbf{0.43 ± 0.04} & \textbf{0.45 ± 0.03}   & 0.72 ± 0.06          \\
                          & CI-GNN (Ave)    & \textbf{0.99 ± 0.01} & 0.32 ± 0.09          & 0.33 ± 0.02            & \textbf{0.86 ± 0.06} \\
                          & CI-GNN (Max)    & \textbf{0.99 ± 0.01} & \textbf{0.32 ± 0.01} & 0.33 ± 0.03            & \textbf{0.69 ± 0.01} \\ \bottomrule
\end{tabular}}
\label{fig:Strategy}
\end{table}

\subsection{Ablation Study}
We further use ablation study to investigate the influence of GraphVAE and causal effect estimator.  Specifically, we compare the classification accuracy among the three variants of CI-GNN including the original model (CI-GNN), CI-GNN-NoCausal (CI-GNN without causal effect estimator), CI-GNN-NonMI (CI-GNN without loss to $I\left ( \alpha; \beta \right )$)  and GIN/GCN (CI-GNN without GraphVAE and causal effect estimator) on four datasets. As shown in Figure~\ref{fig:Ablation}, we observe that CI-GNN outperforms CI-GNN-NonCausal on all datasets, particularly three brain disease datasets, suggesting that the improvements can be attributed to the identifying of causal factors $\alpha$. The more superior performance of CI-GNN than CI-GNN-NonMI, suggesting the effectiveness of independence between $\alpha$ and $\beta$. In addition, the more superior performance of CI-GNN-NonCausal than GIN indicates that GraphVAE is effective and important.

\begin{figure}[ht]
  \centering
  \subfloat[BA2Motif]{\includegraphics[scale=0.55]{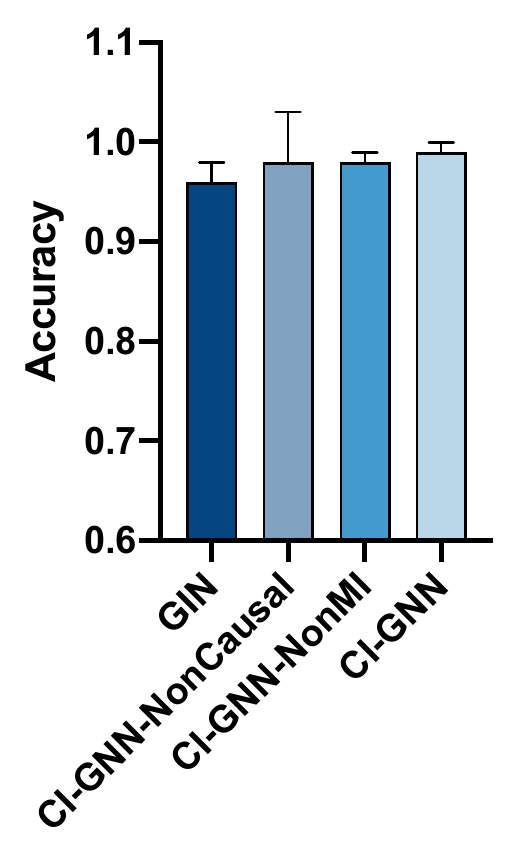}}
  \hfil
  \subfloat[ASD dataset]{\includegraphics[scale=0.55]{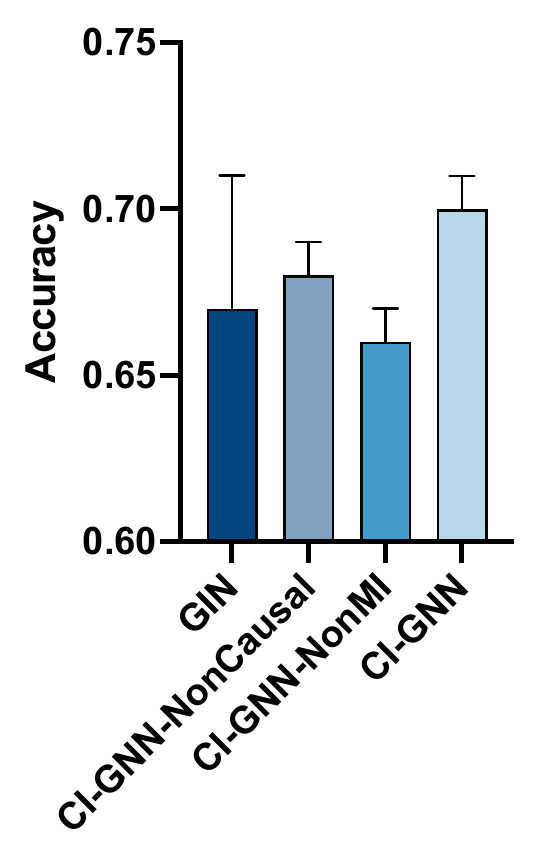}}
  \hfil
  \subfloat[MDD dataset]{\includegraphics[scale=0.55]{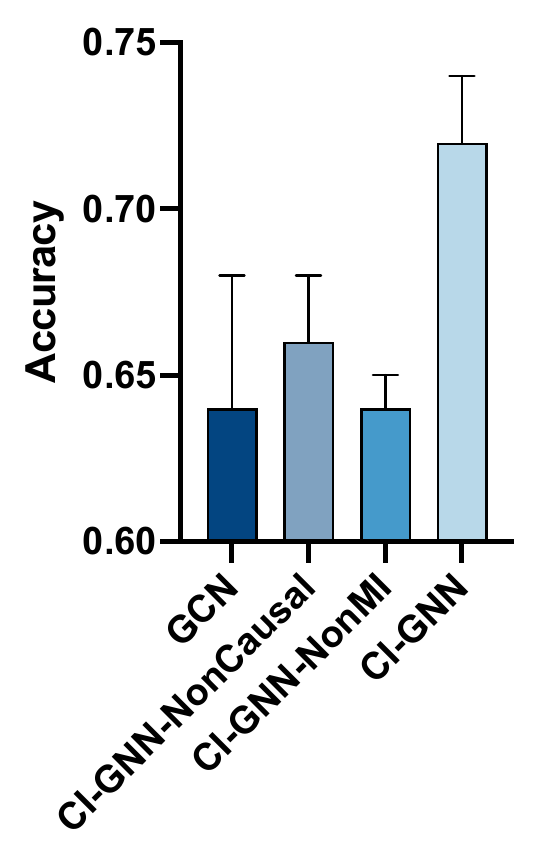}}
  \hfil
  \subfloat[SRPBS dataset]{\includegraphics[scale=0.55]{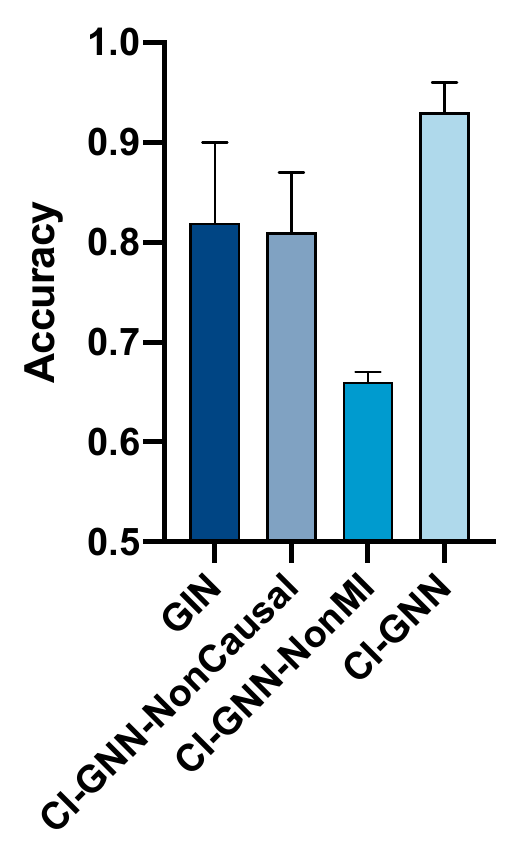}}
  \hfil
  \caption{ Ablation study on four datasets.}
  \label{fig:Ablation}
\end{figure}

To validate that $I\left ( \alpha , \beta \right )$ force the independence between $\alpha$ and $\beta$, we compare the change in correlation between $\alpha$ and $\beta$ before and after using $I\left ( \alpha , \beta \right )$ as loss (CI-GNN and CI-GNN-nonMI). Specifically, we use a popular correlation method Hilbert-Schmidt independence criterion (HSIC)~\cite{gretton2005measuring} to measure the independence between $\alpha$ and $\beta$ in BA-2motif dataset. The according results are reported in the table~\ref{tab:MI}. As can be seen, when using loss $I\left ( \alpha , \beta \right )$, correlation between $\alpha$ and $\beta$ decreases as epochs increase, while correlation remains the same without loss $I\left ( \alpha , \beta \right )$.

\begin{table}[ht!]
\centering
\caption{Changes of Hilbert-Schmidt independence criterion (HSIC) between $\alpha$ and $\beta$ with/without loss $I\left ( \alpha , \beta \right )$. }
\begin{threeparttable}
\renewcommand\arraystretch{1}
\begin{tabular}{@{}c|cccc@{}}
\toprule
\textbf{Epoches} & 1 & 50 & 100 \\ \midrule
CI-GNN       & 0.044          & 0.033        & 0.032         \\
CI-GNN-nonMI & 0.044       & 0.043         & 0.043        \\
\bottomrule
\end{tabular}
\end{threeparttable}
\label{tab:MI}
\end{table}

\subsection{Computational Complexity of CI-GNN}
We compare the efficiency of CI-GNN with three SOTA GNN explainers and report the time required per instance for each model in Table~\ref{tab:Efficiency}. As can we see, the time complexity of CI-GNN is higher than GNNExplainer and PGExplainer due to the complexity of causal effect estimator and training strategy. However, CI-GNN has less computational cost compared with RC-Explainer, indicating that the time spent for CI-GNN is acceptable for graph causal explanation.

\begin{table}[ht!]
\centering
\caption{Efficiency studies of different methods on three graph classification datasets (inference time per instance (ms)). }
\begin{threeparttable}
\renewcommand\arraystretch{1}
\setlength{\tabcolsep}{0.5mm}{}{\begin{tabular}{@{}c|cccc@{}}
\toprule
\textbf{Datasets} & GNNExp & PGExp & RC-Exp & CI-GNN \\ \midrule
BA-2Motifs        & 6.5          & 11.8        & 40.5         & 37.2   \\
ABIDE             & 4.1          & 9.5         & 37.3         & 32.8   \\
REST-meta-MDD     & 1.9          & 9.4         & 37           & 32.8   \\ \bottomrule
\end{tabular}}
\end{threeparttable}
\label{tab:Efficiency}
\end{table}

\subsection{Evaluation on Quality of Explanations}
To assess the interpretation of CI-GNN, we further investigate explanations for graph classification task on BA2Motif and REST-meta-MDD datasets. Figure~\ref{fig:Motif}-\ref{fig:ASD} show the interpretation of BA2Motif, REST-meta-MDD and ABIDE datasets, respectively. 

For BA2Motif dataset, \emph{Cycle} and \emph{House} motifs are causal factors which determine the graph label, while \emph{Tree} motif is non-causal factor which is spuriously associated with the true label~\cite{luo2020parameterized}. In Figure~\ref{fig:Motif}, CI-GNN could successfully recognize the \emph{Cycle} and \emph{House} motifs that explain the labels, while GNNExplainer, PGExplainer and RC-Explainer assign the larger weights on edges out of \emph{Cycle} and \emph{House} motifs, suggesting that \emph{Tree} motif (the spurious correlation) obtained by GNNExplainer, PGExplainer and RC-Explainer could lead to unreliable prediction. 

\begin{figure*}[htbp]
\centering
\includegraphics[scale=0.44]{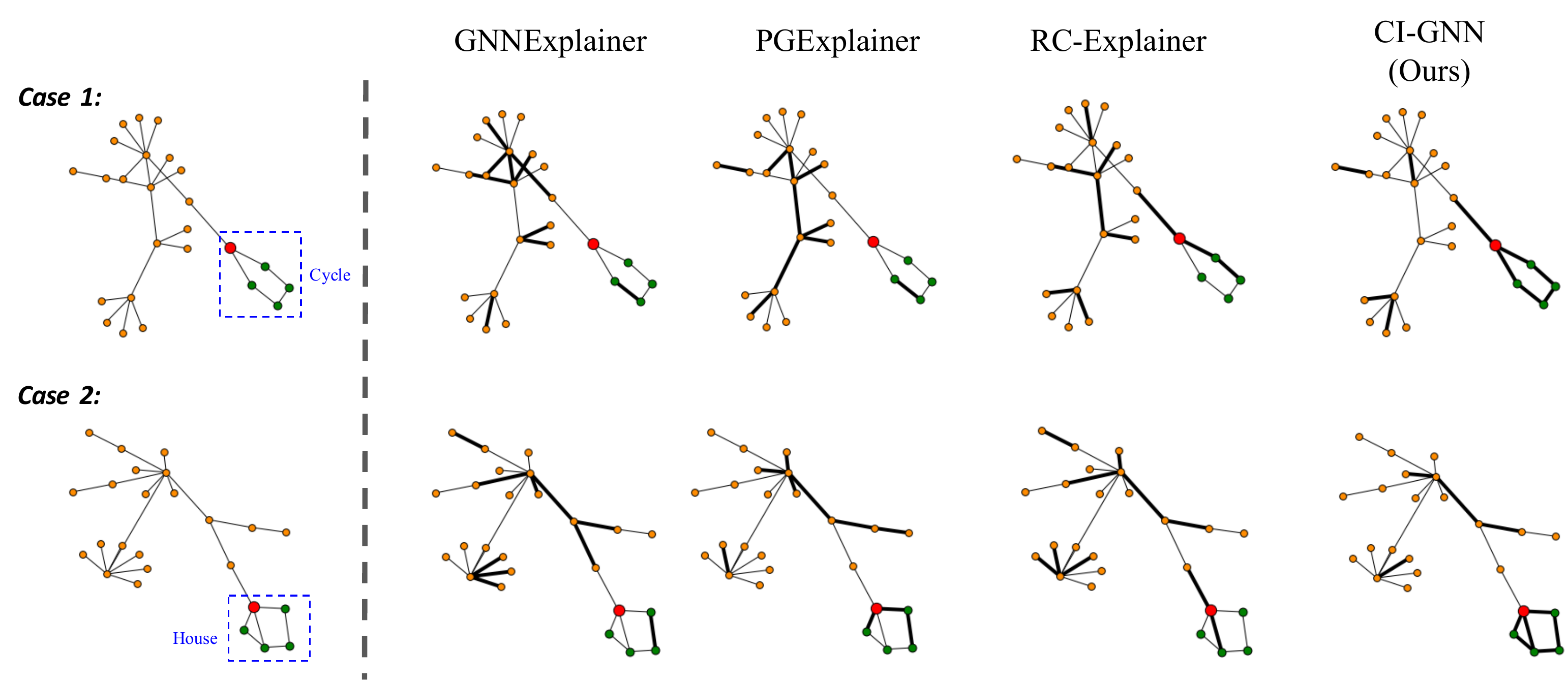}
\caption{Interpretation comparisons on BA2Motif. The \emph{House} and  \emph{Cycle} constructed by the green nodes and red nodes are the ground-truth motifs that determine the graph labels. The orange nodes construct the \emph{Tree} motif which is spuriously associated with the label. The red node connects the ground-truth motif and \emph{Tree} motif. Important edges of explanatory subgraph are highlighted for each model. CI-GNN can identify \emph{House} and \emph{Cycle} motifs, while other methods assign the larger weights on \emph{Tree} motif.}

\label{fig:Motif}
\end{figure*}

We further quantitatively evaluate the explanation accuracy of CI-GNN and other GNN explainers and use $\mu$ to control the size of explanatory subgraph. Specifically, the size of explanatory subgraph can be represented as $\left [ \mu \times \left |  \mathcal{G} \right |   \right ]$. Moreover, we also compute the area under curve (AUC) for the accuracy curve over different $\mu\in \left [ 0.1,0.2, ...1.0 \right ]$. Because the ground truth causal subgraph for real-world brain disease datasets are not available, the explanation performance is compared only on the synthetic dataset. The corresponding results are presented in Table~\ref{tab:explanation}. We observe that CI-GNN can achieve the best explanation performance no matter size of explanatory subgraph, indicating the explanatory reliability of CI-GNN.

\begin{table}[ht!]
\centering
\renewcommand\arraystretch{1}
\caption{Explanation accuracy in terms of AUC on BA2Motif. The best and second best performances are in bold and underlined, respectively.}
\begin{threeparttable}
\begin{tabular}{@{}c|ccc|c@{}}
\toprule
\multicolumn{1}{l|}{$\mu$} & 3             & 4             & 5             & \multicolumn{1}{l}{AUC} \\ \midrule
GNNExplainer           & 0.65          & 0.85          & 0.94          & 0.85                    \\
PGExplainer            & {\ul 0.65}    & {\ul 0.87}    & {\ul 0.95}    & {\ul 0.86}              \\
RC-Explainer           & 0.63          & 0.85          & 0.93          & 0.82                    \\
OrphicX           & 0.67          & 0.83          & \textbf{0.96}          & 0.83                    \\
CI-GNN (Ours)          & \textbf{0.68} & \textbf{0.91} & \textbf{0.96} & \textbf{0.88}           \\ \bottomrule
\end{tabular}
\end{threeparttable}
\label{tab:explanation}
\end{table}

Figure~\ref{fig:MDD} demonstrates the comparison of connectomes for three SOTA graph explainers and CI-GNN on REST-meta-MDD dataset, where the color of each node depends on which the brain network it belongs to. In addition, the size of nodes is determined by the number of connections for them.  It can be seen that the connections of left rectus in CI-GNN are stronger than that of other SOTA graph explainers. This result is consistent with previous clinical finding~\cite{kong2021spatio}, in which left rectus is the most discriminating region for treatment response prediction of MDD. CI-GNN also can identify the interaction between left amygdala and left superior medial frontal gyrus, left middle frontal gyrus, which exhibit significantly abnormalities in patients with MDD as demonstrated by a case-control study~\cite{tassone2022contrasting}. However,  other methods fail to identify these connections.} In addition, we observe that patients with MDD exhibits tight connections between and within cerebellum (nodes with color \textcolor{PineGreen}{pine green}) in three SOTA graph explainers, while these interactions in CI-GNN are much more sparse. As we know, MDD is a common psychiatric disorder involving in affective and cognitive impairments~\cite{belmaker2008major}, but cerebellum is a brain structure directly related to motor function. The hyper-connectivity of the cerebellum in patients with MDD is incompatible with our common sense, suggesting three SOTA graph explainers recognize the spurious correlation. 

In Figure~\ref{fig:ASD}, we show interpretation of GNNExplainer, PGExplainer, RC-Explainer and CI-GNN on ABIDE. As can be seen, CI-GNN is able to recognize the tight connections between somatomotor network (SMN). This finding is in line with the previous study~\cite{lin2022reconfiguration}, where the hidden markov model (HMM) states in ASD patients are characterized by the activation pattern of SMN.  In particular, CI-GNN  can identify interaction between superior temporal gyrus and parahippocampal gyrus. This result is consistent with previous study, in which enhanced FC between superior temporal gyrus and parahippocampal gyrus are observed in patients with ASD~\cite{zhu2022altered}. However, other methods fail to identify some connections associated with ASD. In addition, CI-GNN successfully identifies connections between the visual network and the cerebellum in the ABIDE dataset, a phenomenon not previously reported in the existing literature. This novel discovery may suggest the presence of distinctive neural patterns or interactions that have not been explored in prior studies. Further investigation is warranted to elucidate the potential significance of these identified connections and their implications for our understanding of brain function in the context of psychiatric disorders.

\begin{figure*}[ht!]
\centering
\subfloat[GNNExplainer]{\includegraphics[scale=0.4]{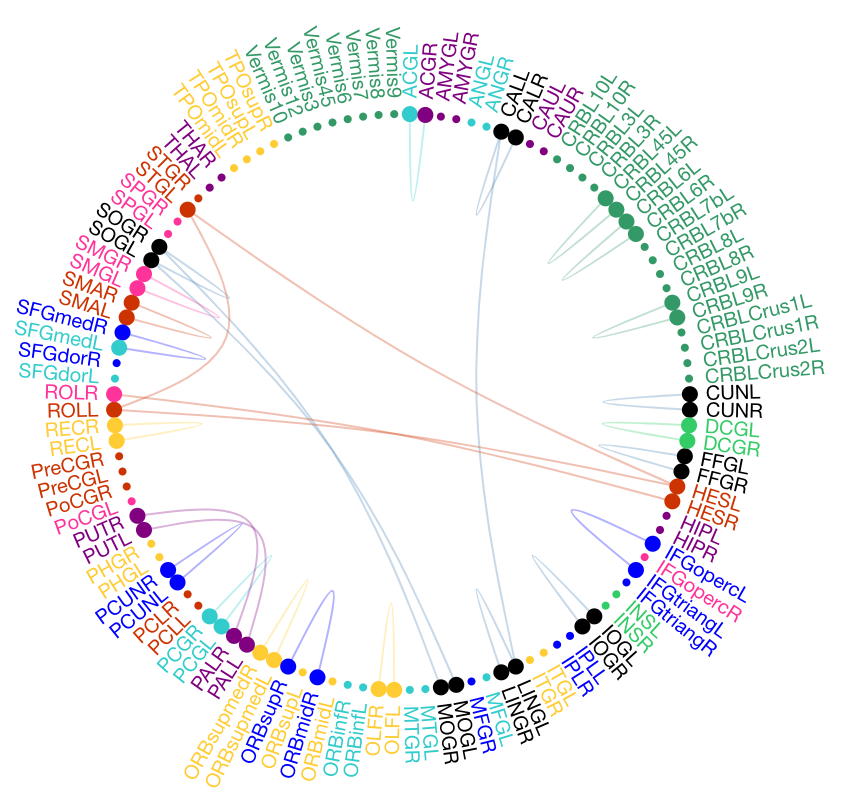}}\hfil
\subfloat[PGExplainer]{\includegraphics[scale=0.4]{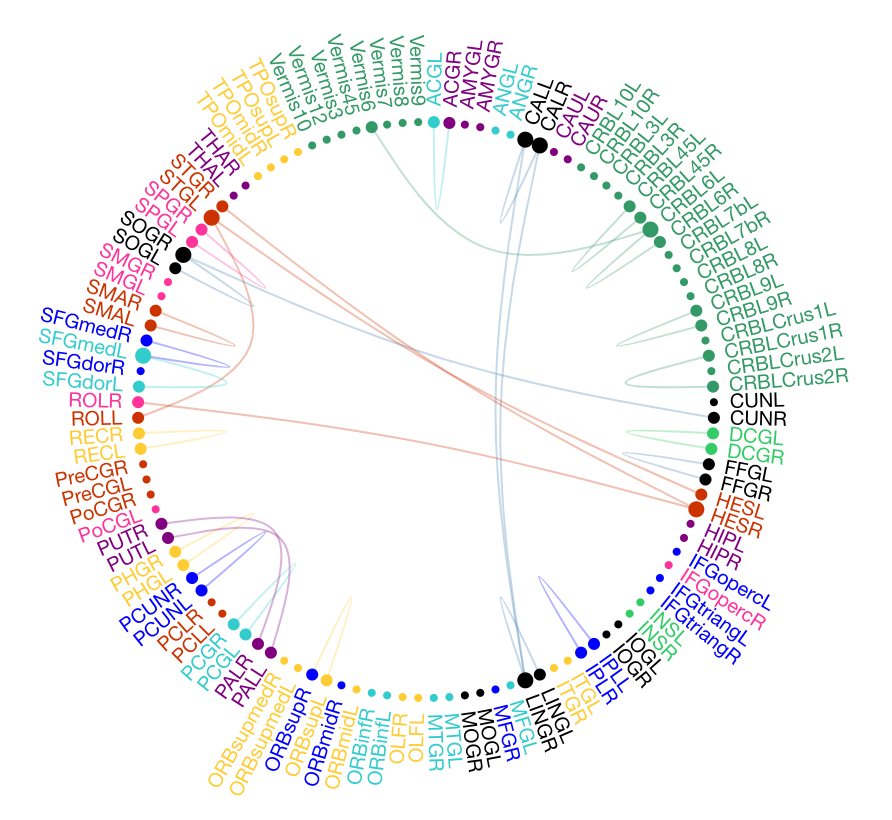}}\hfil
\subfloat[RC-Explainer]{\includegraphics[scale=0.38]{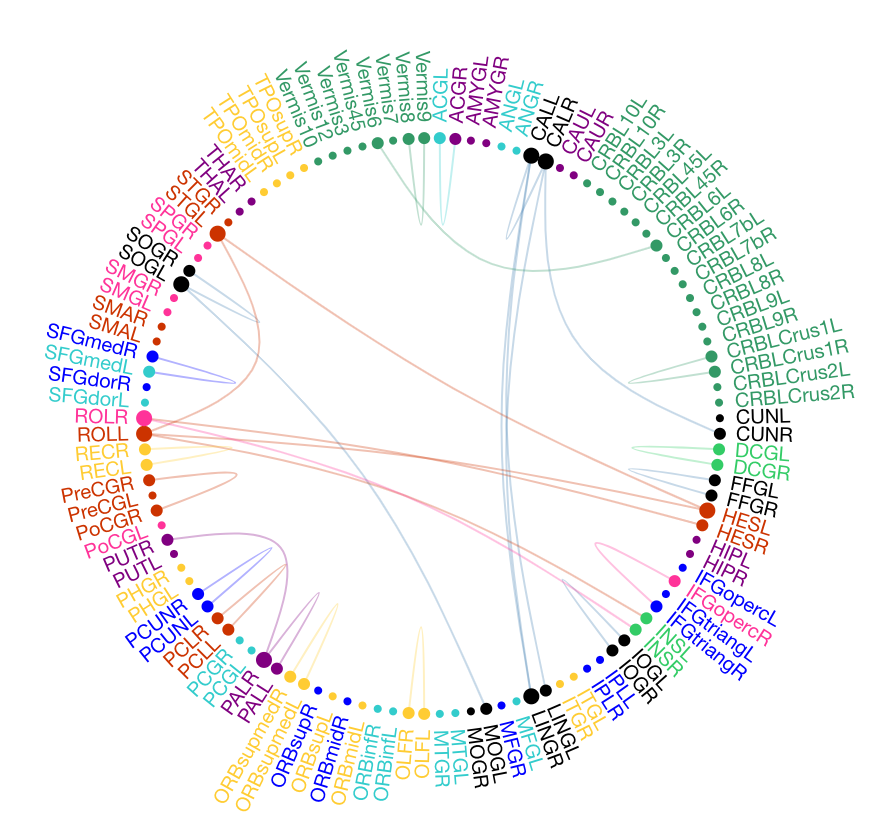}}\hfil
\subfloat[CI-GNN]{\includegraphics[scale=0.38]{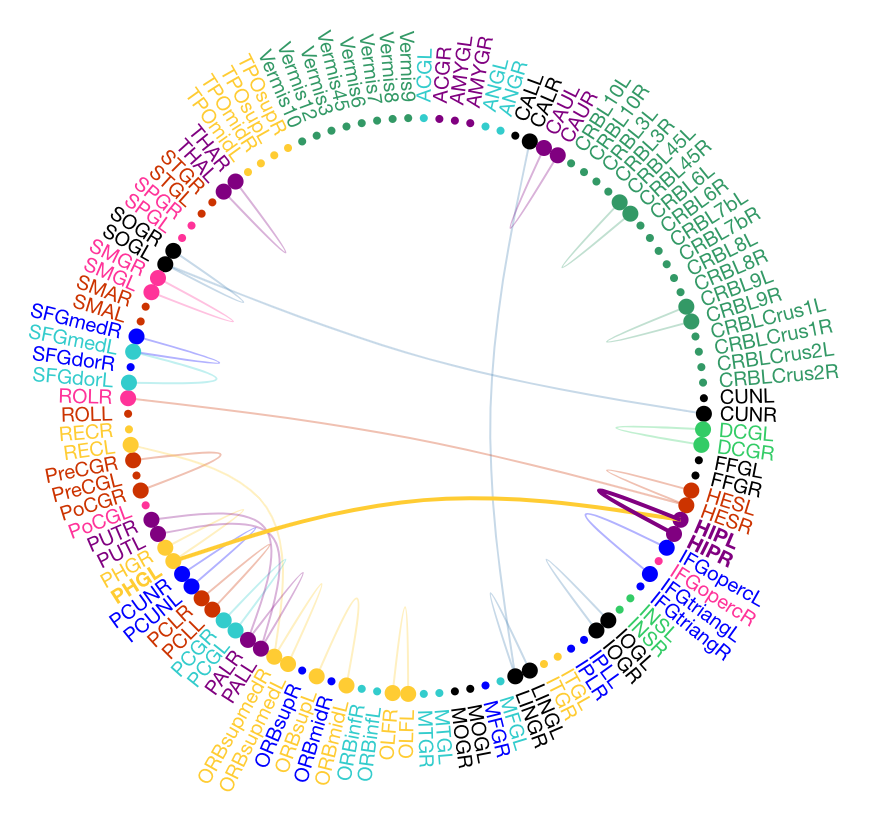}}
\hfil
\caption{ Interpretation comparisons on SRPBS datasets. The colors of brain neural systems are described as: visual network (VN), somatomotor network (SMN), dorsal attention network (DAN), ventral attention network (VAN), limbic network (LIN), frontoparietal network (FPN), default mode network (DMN), cerebellum (CBL) and subcortial network (SBN), respectively. CI-GNN can successfully recognize the interaction between left Hippocampus and left ParaHippocampal that is an intermediate biologic phenotype related to increased genetic risk for schizophrenia. However, other methods fail to identify interaction between Hippocampus and ParaHippocampal.}
\label{fig:SCH}
\end{figure*}

 Figure~\ref{fig:SCH} demonstrates the comparison of connectomes for three SOTA graph
explainers and CI-GNN on SRPBS dataset. We observe that CI-GNN can successfully recognize the interaction between left Hippocampus and left ParaHippocampal that is an intermediate biologic phenotype related to increased genetic risk for schizophrenia~\cite{rasetti2014altered}. However, other methods fail to identify interaction between Hippocampus and ParaHippocampal.

\begin{figure*}[ht!]
\centering
\subfloat[Patient 1]{\includegraphics[scale=0.29]{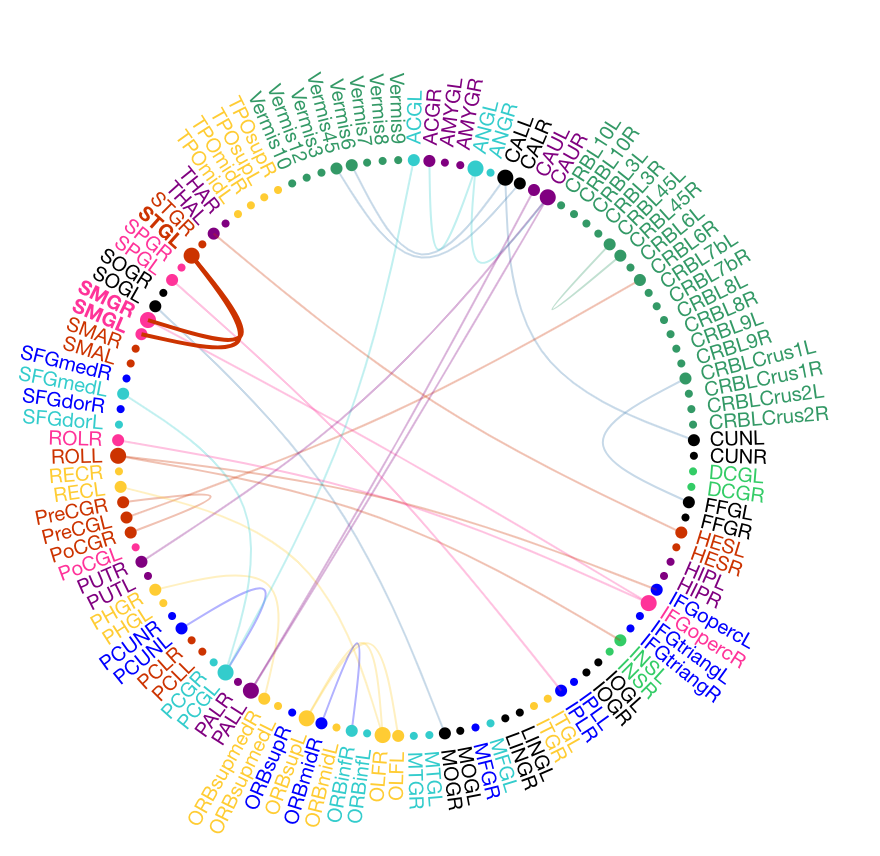}}\hfil
\subfloat[Patient 2]{\includegraphics[scale=0.29]{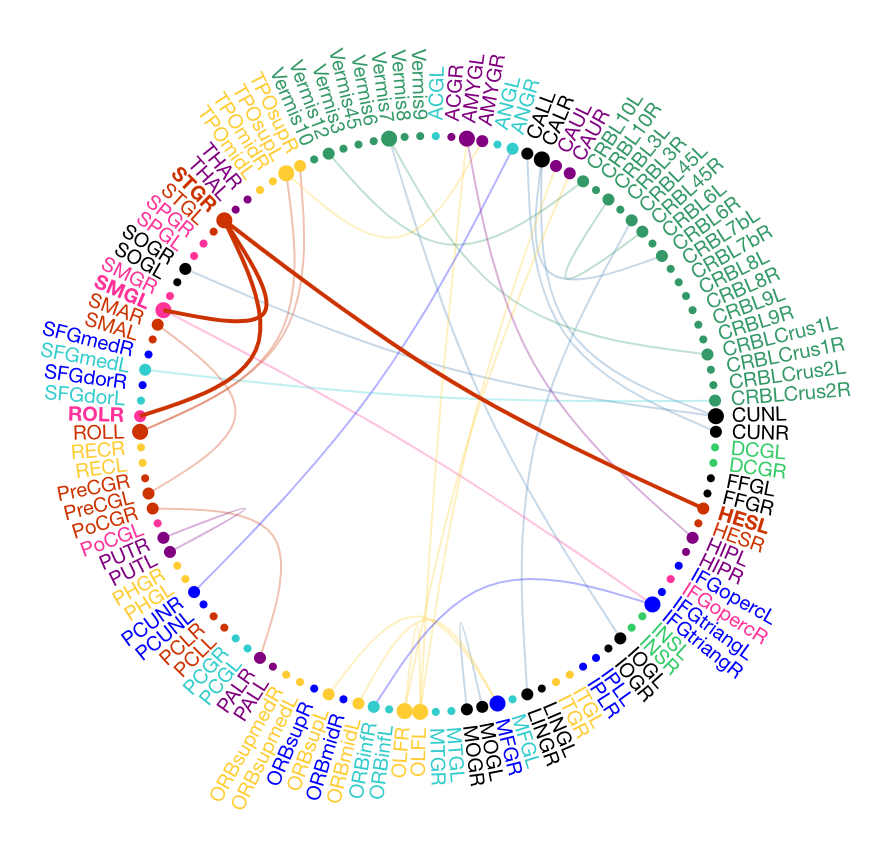}}\hfil
\subfloat[Group Average]{\includegraphics[scale=0.29]{ASD_brain.png}}
\hfil
\caption{ Interpretation comparisons between the group and individual levels on ABIDE datasets. The colors of brain neural systems are described as: visual network (VN), somatomotor network (SMN), dorsal attention network (DAN), ventral attention network (VAN), limbic network (LIN), frontoparietal network (FPN), default mode network (DMN), cerebellum (CBL) and subcortial network (SBN), respectively.}
\label{fig:ABIDE_level}
\end{figure*}

 In figure~\ref{fig:ABIDE_level}, we compare brain patterns of two individual patients with the group average on the ABIDE dataset. As can be seen, both individual subjects and the group average are able to recognize the tight connections within the somatomotor network (SMN), but the spatial patterns between them are not identical.
 
\renewcommand\floatpagefraction{.9}
\begin{figure*}[ht!]
\centering
\subfloat[GNNExplainer]{\includegraphics[scale=0.4]{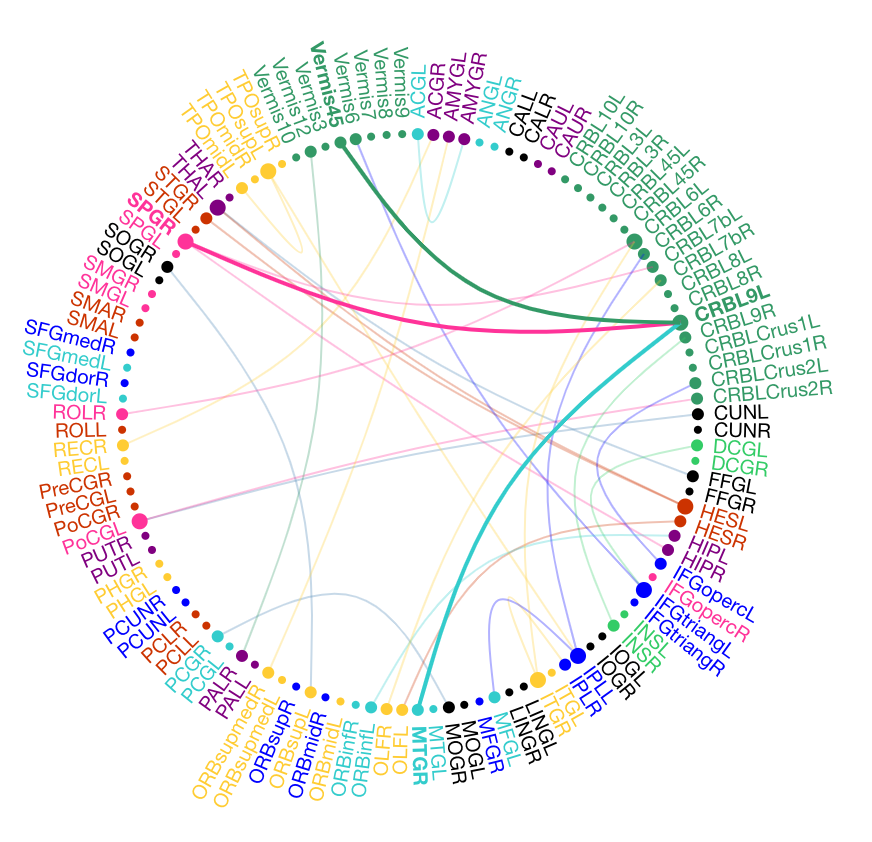}}\hfil
\subfloat[PGExplainer]{\includegraphics[scale=0.4]{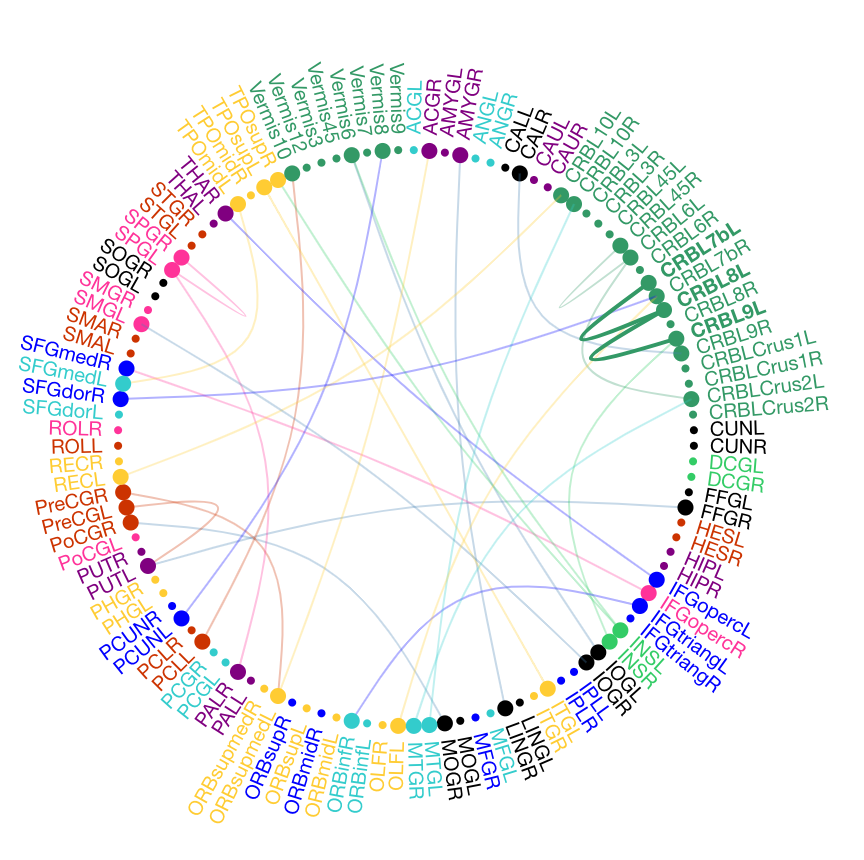}}\hfil
\subfloat[RC-Explainer]{\includegraphics[scale=0.38]{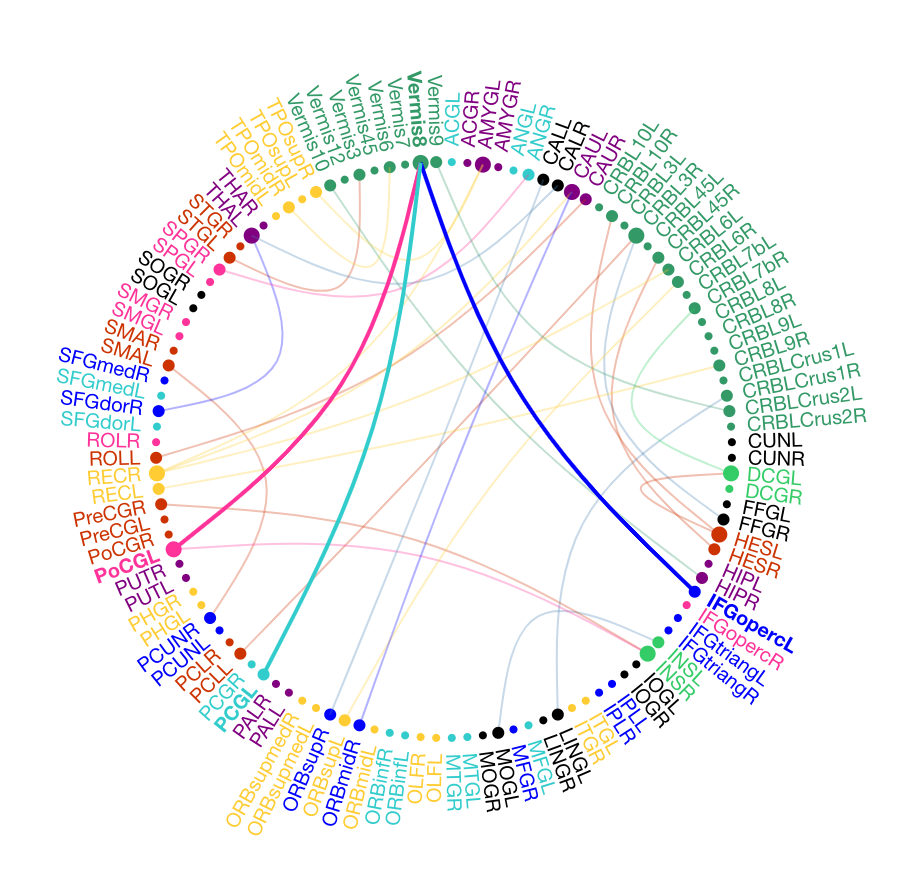}}\hfil
\subfloat[CI-GNN]{\includegraphics[scale=0.38]{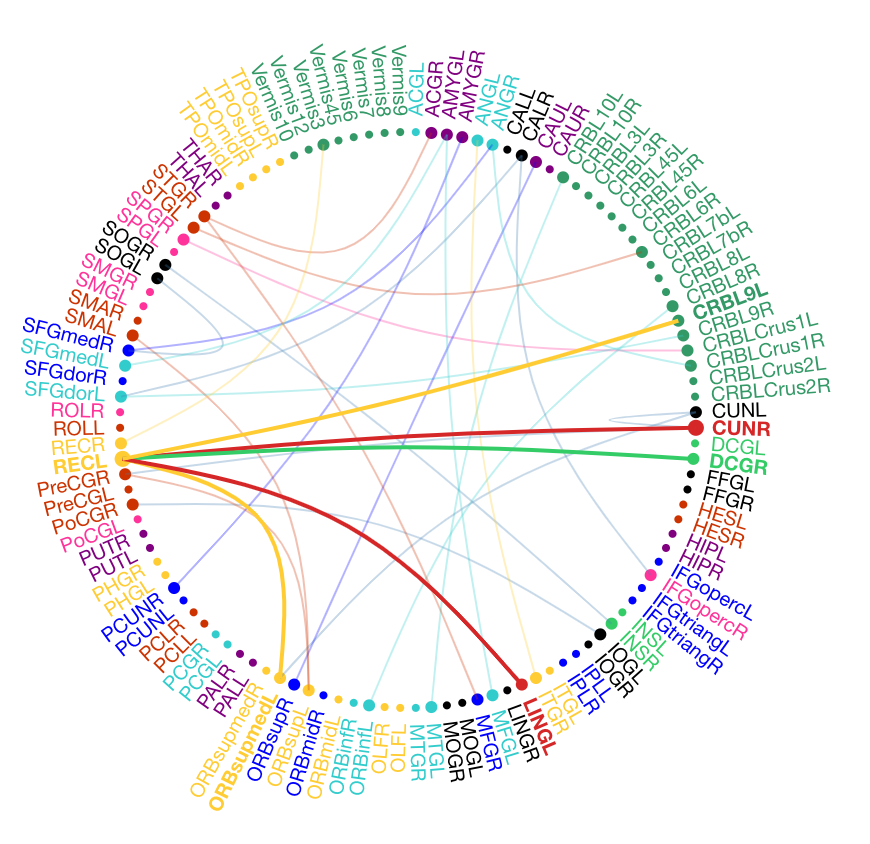}}
\hfil
\caption{Interpretation comparisons on MDD datasets. The colors of brain neural systems are described as: visual network (\color{black}{VN}), somatomotor network({\color{Mahogany}{SMN}}), dorsal attention network ({\color{RubineRed}{DAN}}), ventral attention network ({\color{green}{VAN}}), limbic network ({\color{yellow}{LIN}}), frontoparietal network ({\color{blue}{FPN}}), default mode network ({\color{SkyBlue}{DMN}}), cerebellum ({\color{PineGreen}{CBL}}) and subcortial network ({\color{Orchid}{SBN}}), respectively.  CI-GNN can successfully recognize the interactions of left rectus, which is the most discriminating region for treatment response prediction of MDD. CI-GNN also can identify the interaction between left amygdala and left superior medial frontal gyrus, left middle frontal gyrus, which exhibit significantly abnormalities in patients with MDD as demonstrated by a case-control study. However, other methods mainly find the cerebellum connections and almost fail to identify some connections associated with depression.}
\label{fig:MDD}
\end{figure*}

\renewcommand\floatpagefraction{.9}
\begin{figure*}[ht!]
\centering
\subfloat[GNNExplainer]{\includegraphics[scale=0.4]{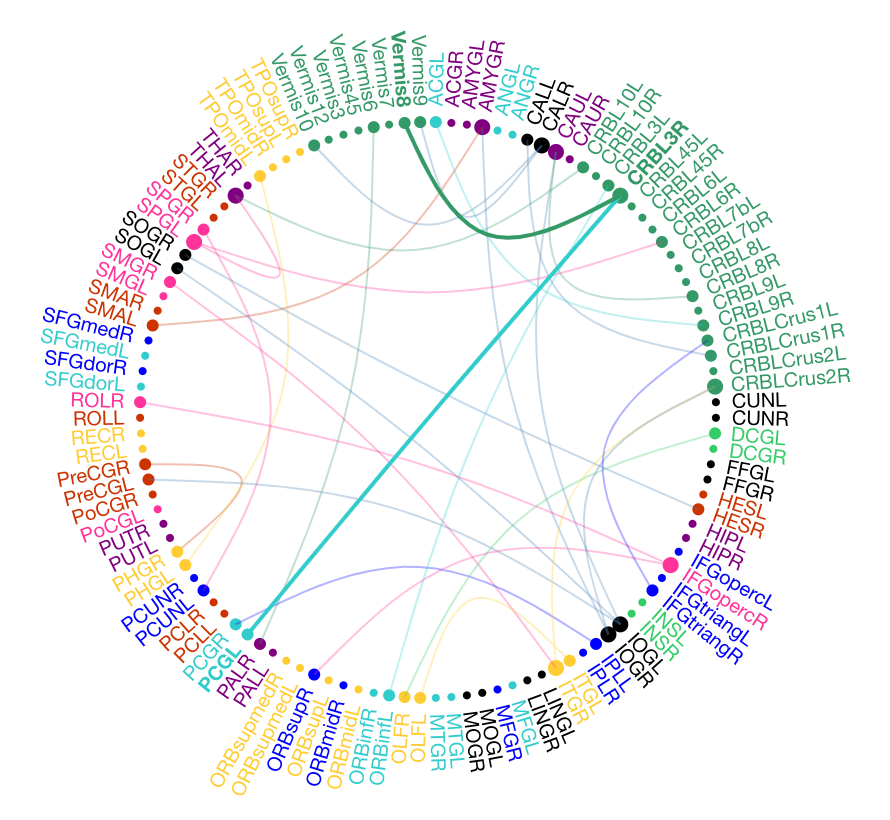}}\hfil
\subfloat[PGExplainer]{\includegraphics[scale=0.4]{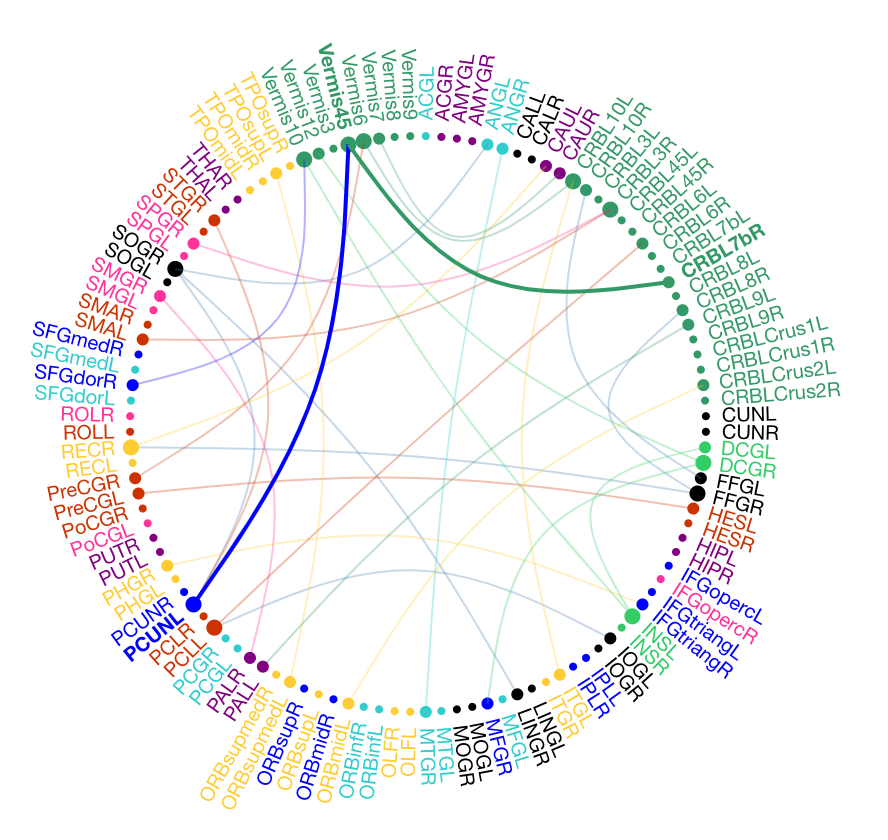}}\hfil
\subfloat[RC-Explainer]{\includegraphics[scale=0.38]{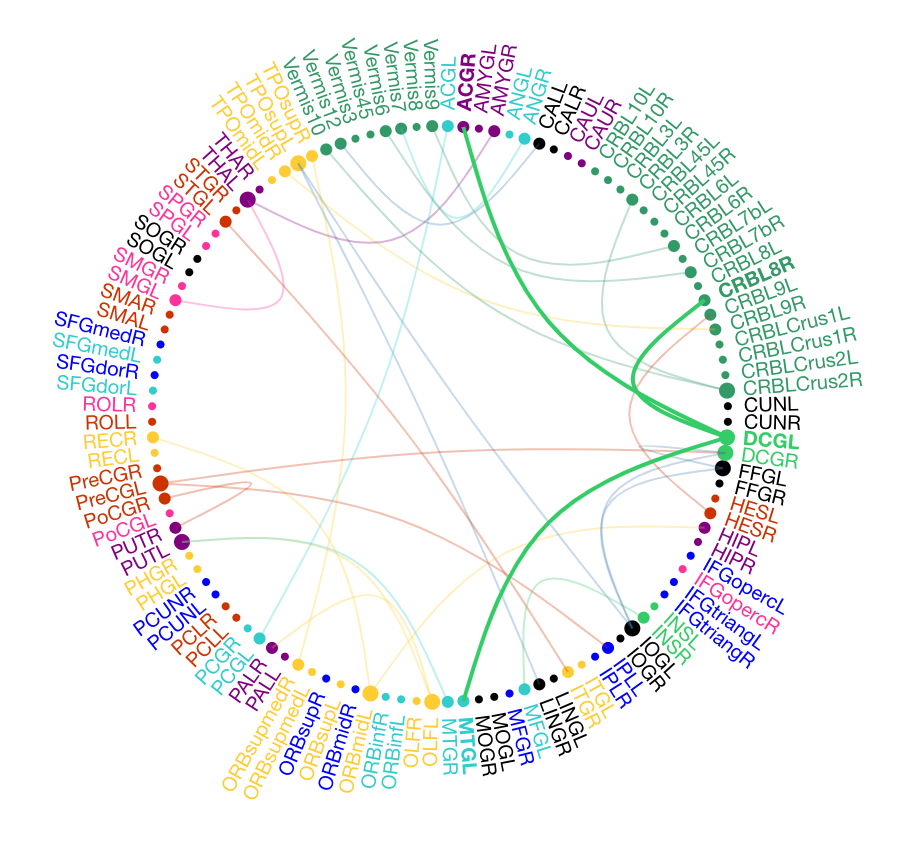}}\hfil
\subfloat[CI-GNN]{\includegraphics[scale=0.38]{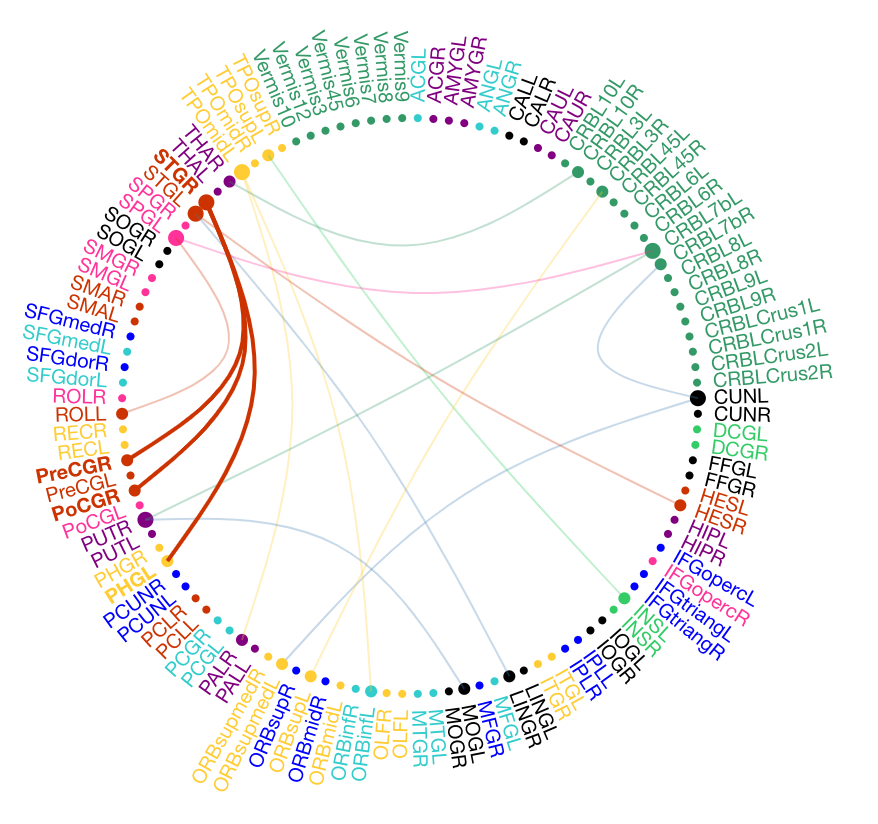}}
\hfil
\caption{Interpretation comparisons on ASD datasets. The colors of brain neural systems are described as: visual network (\color{black}{VN}), somatomotor network({\color{Mahogany}{SMN}}), dorsal attention network ({\color{RubineRed}{DAN}}), ventral attention network ({\color{green}{VAN}}), limbic network ({\color{yellow}{LIN}}), frontoparietal network ({\color{blue}{FPN}}), default mode network ({\color{SkyBlue}{DMN}}), cerebellum ({\color{PineGreen}{CBL}}) and subcortial network ({\color{Orchid}{SBN}}), respectively. CI-GNN can successfully recognize the interactions between SMN that is the important characterization in patients with ASD. In particular, CI-GNN  can identify interaction between superior temporal gyrus and parahippocampal gyrus. This result is consistent with previous study, in which enhanced FC between superior temporal gyrus and parahippocampal gyrus are observed in patients with ASD. However, other methods almost fail to identify some connections associated with autism.}
\label{fig:ASD}
\end{figure*}

\renewcommand\floatpagefraction{.9}

\renewcommand\floatpagefraction{.9}

\subsection{Evaluation on Other Graph Datasets}
To assess the robustness and generalization of our model, we further test the performance and interpretation of three bioinformatics datasets including MUTAG, PROTEINS and NCI1. The data statistics of three bioinformatics datasets are provided in the Appendix. Table~\ref{tab:Performance1} demonstrates the performance results on MUTAG, PROTEINS and NCI1. As can be seen, CI-GNN achieves the highest performance in accuracy, F1-score and MCC, which indicates that our CI-GNN can also be generalized to molecular data (rather than just brain networks). Ablation studies are provided in the Appendix.
\begin{table*}[ht!]
\centering
\caption{The classification performance of accuracy, F1 and MCC and standard deviations of CI-GNN and the baselines on three bioinformatics datasets. The best and second best performances are in bold and underlined, respectively.}
\begin{threeparttable}
\renewcommand\arraystretch{1}
\resizebox{\linewidth}{!}{\begin{tabular}{@{}c|ccc|ccc|ccc@{}}
\toprule
\multirow{2}{*}{Methods} & \multicolumn{3}{c|}{MUTAG}                                                             & \multicolumn{3}{c|}{PROTEINS}                                                          & \multicolumn{3}{c}{NCI1}                                           \\ \cmidrule(l){2-10} 
                         & Accuracy                                 & F1                   & MCC                  & Accuracy                                 & F1                   & MCC                  & Accuracy             & F1                   & MCC                  \\ \midrule
GCN                      & 0.87 ± 0.08                              & 0.90 ± 0.06          & 0.90 ± 0.06          & 0.76 ± 0.03                              & 0.66 ± 0.05          & 0.49 ± 0.05          & 0.76 ± 0.01          & {\ul 0.75 ± 0.03}          & 0.52 ± 0.01          \\
GAT                      & {\ul 0.90 ± 0.05}                        & {\ul 0.93 ± 0.04}    & {\ul 0.93 ± 0.04}    & 0.78 ± 0.03                              & 0.63 ± 0.12          & 0.50 ± 0.12          & 0.72 ± 0.01          & 0.71 ± 0.03          & 0.45 ± 0.01          \\
GIN                      & {\ul 0.90 ± 0.06}                        & {\ul 0.93 ± 0.03}    & {\ul 0.93 ± 0.03}    & 0.77 ± 0.03                              & 0.67 ± 0.01          & 0.52 ± 0.05          & {\ul 0.82 ± 0.01}    & \textbf{0.82 ± 0.01} & {\ul 0.63 ± 0.04} \\
SIB                      & 0.75 ± 0.05                              & 0.82 ± 0.02          & 0.82 ± 0.02          & 0.79 ± 0.01                              & 0.72 ± 0.03          & 0.56 ± 0.03          & 0.65 ± 0.01          & 0.69 ± 0.02          & 0.30 ± 0.01          \\
DIR-GNN                      & 0.87 ± 0.08                              & 0.92 ± 0.04          & 0.61 ± 0.24          & 0.77 ± 0.01                              & 0.72 ± 0.02          & 0.58 ± 0.02          & 0.71 ± 0.01          & 0.70 ± 0.01          & 0.42 ± 0.01          \\
GNNExplainer             & 0.88 ± 0.03                              & 0.91 ± 0.01          & 0.91 ± 0.01          & {\ul 0.80 ± 0.01}                        & 0.69 ± 0.01          & 0.55 ± 0.01          & 0.68 ± 0.01          & 0.63 ± 0.05          & 0.60 ± 0.01          \\
PGExplainer              & 0.87 ± 0.03                              & 0.87 ± 0.04          & 0.87 ± 0.04          & {\ul 0.80 ± 0.02}                        & {\ul 0.73 ± 0.04}    & {\ul 0.60 ± 0.04}    & 0.65 ± 0.02          & 0.65 ± 0.03          & 0.30 ± 0.05         \\
RC-Explainer             & 0.82 ± 0.06                              & 0.87 ± 0.03          & 0.87 ± 0.03          & 0.67 ± 0.06                              & 0.42 ± 0.12          & 0.27 ± 0.08          & 0.65 ± 0.03          & 0.65 ± 0.13          & 0.31 ± 0.08          \\
CI-GNN (Ours)            & \multicolumn{1}{l}{\textbf{0.93 ± 0.03}} & \textbf{0.94 ± 0.02} & \textbf{0.94 ± 0.02} & \multicolumn{1}{l}{\textbf{0.82 ± 0.02}} & \textbf{0.76 ± 0.04} & \textbf{0.61 ± 0.05} & \textbf{0.83 ± 0.02} & \textbf{0.82 ± 0.02} & \textbf{0.65 ± 0.03} \\ \bottomrule
\end{tabular}}
\end{threeparttable}
\label{tab:Performance1}
\end{table*}

\begin{figure*}[ht]
\centering
\includegraphics[scale=0.4]{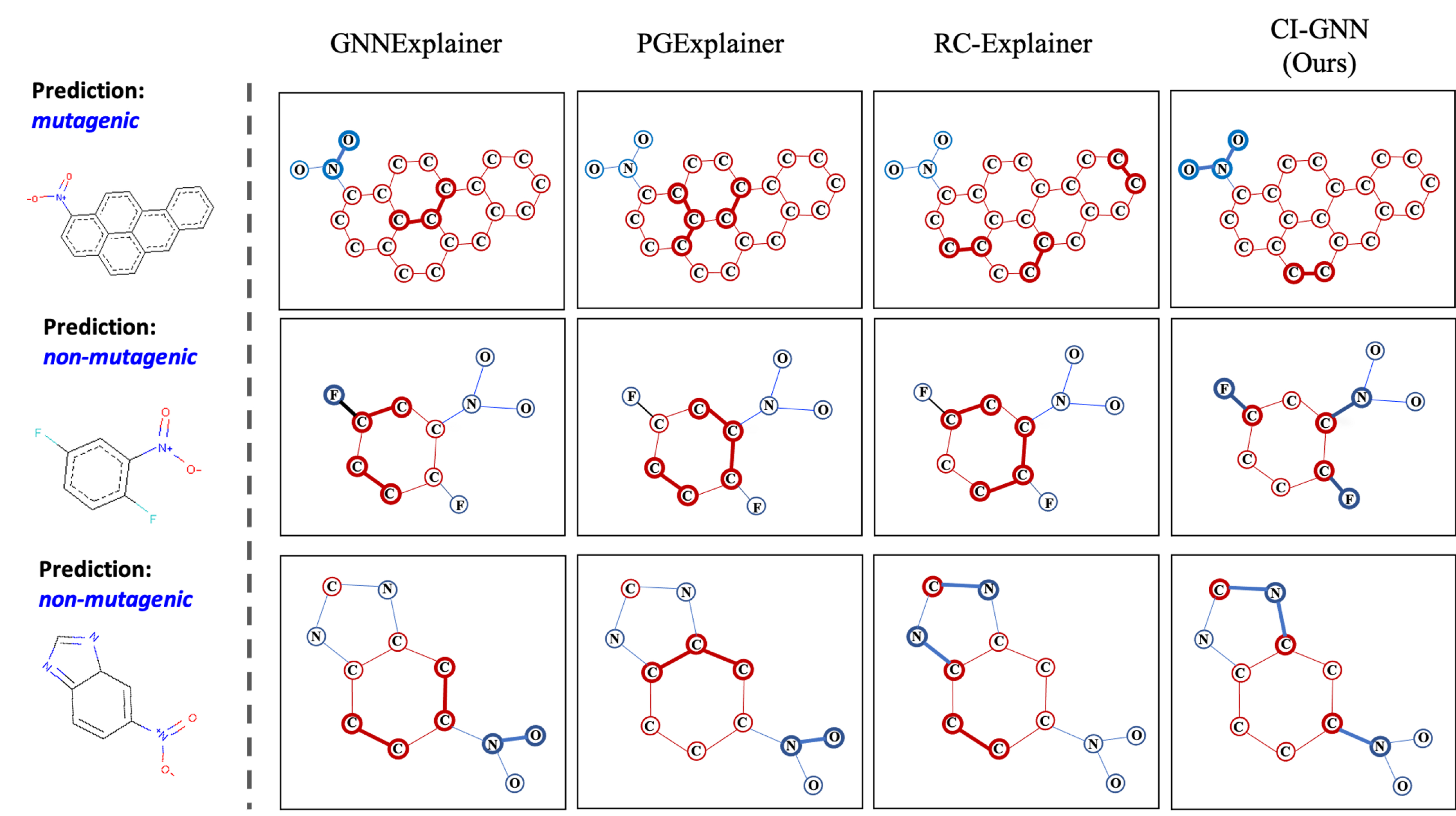}
\caption{Interpretation comparisons on MUTAG. Important edges of explanatory subgraph are highlighted for each model.}
\label{fig:Molecule}
\end{figure*}

Furthermore, we compare the interpretation of CI-GNN between GNNExplainer, PGExplainer and RCExplainer on MUTAG in Figure~\ref{fig:Molecule}. As we know, labels of MUTAG include mutagenic and non-mutagnic. Mutagenic structure is determined by carbon rings and \emph{$NO_{2}$} groups~\cite{debnath1991structure}. Obviously, we observe that CI-GNN correctly identifies \emph{$NO_{2}$} groups but other explainers fail to obtain. Furthermore, CI-GNN also recognizes $F$ atom on carbon rings and $N$ atom in carbon rings that the causal factors for non-mutagic. 

\section{Conclusions and Future Work}

In this work, we proposed the first \emph{built-in} interpretable graph neural networks from a causal perspective, in the sense that  our model is able to learn a disentangled latent representation $\alpha$ (and its associated subgraph structure) that is causally related to final decision $Y$. In terms of application, we evaluated our model on two large-scale, multi-site brain disease datasets and tested its performances on brain network-based psychiatric disorder diagnosis. Our model consistently achieved the highest classification accuracy in both datasets and identified subgraph biomarkers which are coincide with clinical observations. 

In the future, we will evaluate the generalization ability of our model to out-of-distribution (OOD) datasets. This is just because our model is able to extract a causal latent representation that is also possibly ``invariant" under distributional shift. 
We will also test the performances of other possible causal effect estimators (e.g., the mutual information term $I( \alpha;Y )$ or $I( \alpha;\hat{Y} )$) and theoretical investigate their inner connections.   In addition, we will incorporate human/clinicians expert knowledge into the design of GNNs, which could further enhance the performance of GNNs.



\section{Acknowledgments}
This work was supported by the National Natural Science Foundation of China with grant numbers (U21A20485, 62088102, 61976175). 

\vfill
\newpage
\appendix
\begin{spacing}{2.0}
{\raggedright\huge\bfseries\centering{\appendixpagename}}
\end{spacing}

This document contains the Appendix for the \textit{``CI-GNN: A Causality-Inspired Graph Neural Network for Interpretable Brain Network Analysis"} manuscript. It is organized into the following topics and sections:

\begin{enumerate}
\item Proof of Corollary~1
\item Detailed Experimental Setup and Additional Experimental Results
\begin{enumerate}
\item Preprocessing of Two Brain Disease Datasets
\item Data statistics of three bioinformatics datasets
\item Architecture of CI-GNN
\item Evaluation Metrics
\item Additional Experimental Results for Performance
\item Leave-One-Site-Out Cross Validation
\item {  Inner connections between different causal effect estimators including $I(\alpha;Y|\beta)$ and $I(\alpha;\hat{Y}|\beta)$}
\end{enumerate}
\item Minimal Implementation of CI-GNN in PyTorch
\end{enumerate}

\section{Proof of Corollary~1}
\noindent\textbf{Corollary~1.} \emph{$I(\alpha;Y|\beta)$ is able to measure the causal effect of $\alpha$ on $Y$ when ``imposing" $\beta$}. \\
\emph{Proof.} From a Granger causality perspective~\cite{granger1969investigating}, variable $X$ causes another variable $Y$ if, in a statistical sense, the prediction of $Y$ is improved by incorporating information about $X$. The general idea of Granger causality can be extended if a third variable $Z$ is taken into account~\cite{chen2006frequency}. In this case, the evaluation of the conditional Granger causality $I\left ( X\to Y|Z  \right )$ in the time domain is fairly straightforward through comparison of two predictions of $Y$, one when $Z$ is given, the other when $\left ( X ,Z \right )$ are given together.

Let $X$ denotes observational random variables (e.g., features) and $Y$ a discrete valued random variable (e.g., class labels) which takes $L$ different vales $\left ( c_{1}, c_{2}, ..., c_{L}  \right )$ . We aim to infer $Y$ from $X$. Suppose the predicted output is $\hat{Y}$, then the prediction error probability from $X$ to $Y$ is $p_{e}\left ( X\to Y \right ) =P\left ( Y\ne \hat{Y}  \right )$. In Bayesian statistics, the upper and lower bounds of $p_{e}$ can be derived by information-theoretic criteria. Specifically, these bounds can be determined as~\cite{ozdenizci2019information}:
\begin{equation}\label{eq:bond}
\begin{aligned}
\frac{H\left ( Y|X \right ) }{2} \ge p_{e} \ge \frac{H\left ( Y|X \right )-1 }{\log{L}},
\end{aligned}
\end{equation}
with $H\left ( Y|X \right )$ denotes the Shannon conditional entropy of $Y$ given $X$. In fact, the lower bound in Eq.~(\ref{eq:bond}) is the famed Fano’s inequality~\cite{fano1961transmission}, whereas the upper bound is also known as the Hellman-Raviv bound~\cite{hellman1970probability}. Together, these inequalities suggest that the prediction performance of $X$ to $Y$ can be quantified by the conditional entropy $H\left ( Y|X \right )$. 

Now, to quantify the strength of causal effect of variable $\alpha$ on $Y$, when imposing on a third variable $\beta$, one can directly compare the difference between prediction error probabilities $p_{e}\left ( \alpha \to Y \right )$  and $p_{e}\left ( \left ( \alpha, \beta \right ) \to Y \right )$ , in a Granger sense (i.e., if the prediction of $Y$ can be improved by incorporating information about $\beta$):
\begin{equation}
\begin{aligned}
I\left ( \alpha \to Y|\beta \right ) =p_{e}\left ( \alpha \to Y \right ) -p_{e}\left ( \left ( \alpha,\beta \right ) \to Y  \right ).
\end{aligned}
\end{equation}

On the other hand, given Eq.~(\ref{eq:bond}), one can replace $p_{e}\left ( \alpha \to Y \right )$ and $p_{e}\left ( \left ( \alpha, \beta \right ) \to Y \right )$ by $H\left(Y|\alpha \right)$ and $H\left(Y|\alpha,\beta \right)$, as a surrogate. Therefore, we obtain:
\begin{equation}
\begin{aligned}
I\left ( \alpha \to Y|\beta \right ) \approx H\left ( Y|\alpha \right )-H\left ( Y|\alpha,\beta \right )=I\left (Y; \alpha|\beta \right ).
\end{aligned}
\end{equation}

\section{Detailed Experimental Setup}

\subsection{Preprocessing of Two Brain Disease Datasets}

Table~\ref{tab:clinical} shows the demographic and clinical characteristics of both brain disease datasets.  For ABIDE, we remove 35 Female healthy controls to reduce gender bias and finally choose 1064 participants in this study. Resting-state fMRI data are preprocessed using the statistical parametric mapping (SPM) software\footnote{\url{https://www.fil.ion.ucl.ac.uk/spm/software/spm12/}}. The
preprocessing procedures include discarding the initial 10 volumes, slice-timing correction, head motion correction, space normalization, smoothing ($6$ mm), temporal bandpass filtering (0.01-0.1 HZ) and regressing out the effects of head motion, white matter and cerebrospinal fluid signals (CSF). For REST-meta-MDD dataset, we choose 1, 604 subjects according to exclusion criteria including poor spatial normalization, bad coverage, excessive head 
movement, centers with fewer than 10 participants and incomplete time series data and information. We employ preprocessed data previously made available by the REST-meta-MDD for further analysis.  The resting-state fMRI data of SRPBS dataset are followed by the same preprocessing procedure of ABIDE dataset.

After preprocessing, we extract mean time series of each region of interest (ROI) according to the automated anatomical labelling (AAL) atlas. Furthermore, we calculate the functional connectivity between all ROI pairs by Pearson’s correlation and further generate 116 $\times$ 116 functional connectivity matrix (brain network).

\begin{table}[ht!]
\caption{Demographic and clinical characteristics.}
\renewcommand\arraystretch{1.5}
\resizebox{\linewidth}{!}{\begin{tabular}{@{}c|cccccc@{}}
\toprule
\multirow{2}{*}{Characteristic} & \multicolumn{2}{c}{ABIDE} & \multicolumn{2}{c}{Rest-meta-MDD} & \multicolumn{2}{c}{SRPBS}    \\ \cmidrule(l){2-7} 
                                & ASD         & TD          & MDD             & HC              & Schizophrenia & HC           \\ \midrule
Sample Size                     & 528         & 536         & 828             & 776             & 92            & 92           \\
Age                             & 17.0 ± 8.4  & 17.2 ﻿± 7.6 & 34.3 ± 11.5     & 34.4 ± 13.0     & 39.6 ± 10.4   & 38.0 ﻿± 12.4 \\
Gender (M/F)                    & 464/64      & 471/65      & 301/527         & 318/458         & 47/45         & 60/32        \\ \bottomrule
\end{tabular}}\label{tab:clinical}
\end{table}

\subsection{Data statistics of three bioinformatics datasets}

In addition to the synthetic dataset and brain disease datasets, we also select three bioinformatics datasets including MUTAG~\cite{debnath1991structure}, PROTEINS~\cite{borgwardt2005protein} and NCI1~\cite{wale2008comparison}:

\begin{itemize}
\item{MUTAG~\cite{debnath1991structure}: It has 288 molecule graphs and 2 graph labels including mutagenic and nonmutagenic.}

\item{PROTEINS~\cite{borgwardt2005protein}: It is a graph classification bioinformatics dataset that includes 1,113 graph structures of proteins. For each graph, nodes mean secondary structure elements (SSEs) and the existence of an edge is determined by whether two nodes are adjacent along the amino-acid sequence or in space.}

\item{NCI1~\cite{wale2008comparison}: It is a bioinformatics dataset with $4,110$ graph-structured chemical compounds. Positive sample is judged by whether it has the ability to suppress or inhibit the growth of a panel of human tumor cell lines}
\end{itemize}

The statistics of three bioinformatics datasets are summarized in Table~\ref{tab:data}.

\begin{table}[ht!]
\centering
\caption{Statistics of three bioinformatics datasets.}
\begin{threeparttable}
\renewcommand\arraystretch{1}
\begin{tabular}{@{}c|cccc@{}}
\toprule
Datasets & \# of Edges (avg) & \# of Nodes (avg) & \# of Graphs & \# of Classes \\ \midrule
MUTAG    & 9.90             & 17.93             & 188          & 2             \\
PROTEINS & 72.8             & 39.06             & 1113         & 2             \\
NCI1     & 32.3              & 29.87             & 4110         & 2             \\ \bottomrule
\end{tabular}
\end{threeparttable}
\label{tab:data}
\end{table}

\subsection{Architecture of CI-GNN}
Table~\ref{tab:architecture} summarizes the architecture of CI-GNN including the neurons of GCN encoder $\phi$,  multi-head decoder $\theta_{1}$, linear decoder $\theta_{2}$ and the basic classifier $\varphi$.  
\begin{table}[ht!]
\centering
\caption{Architecture of CI-GNN.}
\begin{threeparttable}
\renewcommand\arraystretch{1.5}
\setlength{\tabcolsep}{0.5mm}{}{\begin{tabular}{|c|c|}
\hline
$\#$Neurons of GCN Encoder $\phi$       & [128,64]              \\ \hline
$\#$Neurons of Multi-head Decoder $\theta_{1}$ & [16] and [1]                 \\ \hline
$\#$Neurons of Linear Decoder $\theta_{2}$ & [1]                 \\ \hline
$\#$Neurons of Classifier $\varphi$         & [128,128,128,64,32,2] \\ \hline
\end{tabular}}
\end{threeparttable}
\label{tab:architecture}
\end{table}

\subsection{Evaluation Metrics}
We use accuracy, f1-score and matthew’s
cerrelatien ceefficient (MCC) as performance metrics to evaluate the effectiveness of our model. These measures are defined as:

\begin{equation}
\mbox{accuracy} = \frac{\mbox{TP} + \mbox{TN}}{\mbox{TP} + \mbox{TN} + \mbox{FP} + \mbox{FN}},
\end{equation} 




\begin{equation}
\mbox{F1} = \frac{2 \times \mbox{TP}}{2 \times \mbox{TP} + \mbox{FP} + \mbox{FN}},
\end{equation} 

\begin{small}
\begin{equation}
\begin{aligned}
\mbox{MCC}=\frac{\mbox{TP} \! \times \mbox{TN} - \mbox{FP} \! \times \mbox{FN}}{ \left [  \left ( \mbox{TP} \! + \mbox{FP} \right) \! \times \left ( \mbox{TP} \! + \mbox{FN}  \right ) \! \times \left ( \mbox{TN} \! + \mbox{FP}  \right ) \! \times \left ( \mbox{TN} \! + \mbox{FN}  \right )  \right ] ^{0.5}  } ,
\end{aligned}
\end{equation} 
\end{small}
where TP and TN refer to the true-positive and true-negative values, and FP and FN denote false-positive and false-negative values, respectively.

\subsection{Additional Experimental Results for Performance}

\begin{figure}[ht]
\centering
\includegraphics[scale=0.7]{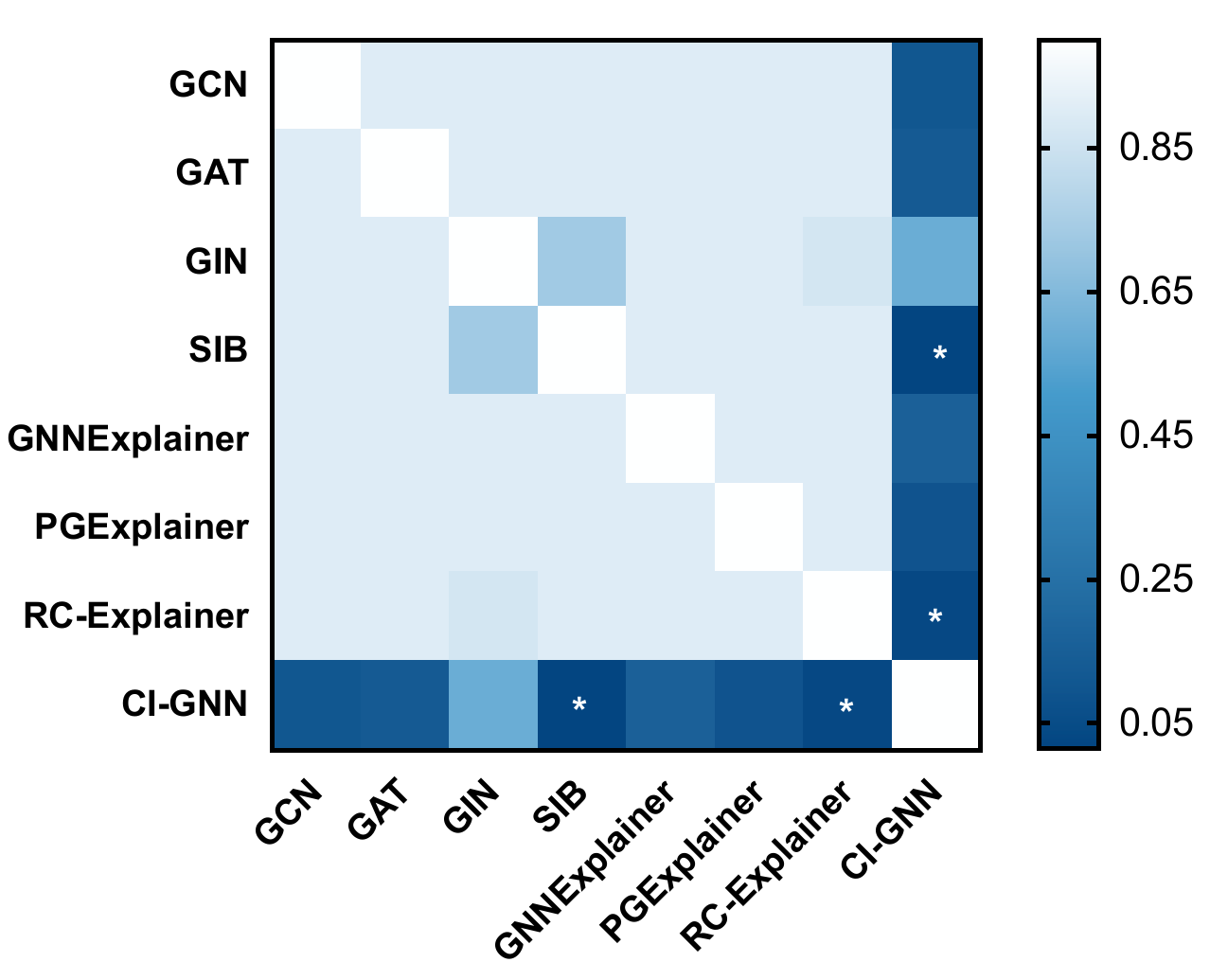}
\caption{The statistical differences ($\mit{p}$-values) between all competing methods pairs using a Nemenyi’s post-hoc test are shown. Significant differences ($\mit{p} <$ 0.05) are marked with a star. Our method are significantly different from the SIB and RC-Explainer.}
\label{fig:test}
\end{figure}

In order to validate the superior performance of CI-GNN, we report the ranks over different performance metrics in each dataset and further perform Nemenyi’s post-hoc test~\cite{nemenyi1963distribution} to test the statistical difference in all methods. According results are shown in Table~\ref{tab:Performance4} and Figure~\ref{fig:test}. Results indicate that CI-GNN  is able to achieve the best performance on all datasets no matter the performance metrics. Furthermore, we observe that CI-GNN are significantly different from the SIB and RC-Explainer.

\begin{table*}[ht!]
\centering
\caption{A summarization of different approaches and their average ranks over different performance metrics in each dataset. The overall average ranks over different datasets are shown. The best two performance in each dataset are in bold and underlined, respectively.}

\begin{threeparttable}
\resizebox{\linewidth}{!}{
\begin{tabular}{@{}c|cccccc|c@{}}
\toprule
Methods      & BA-2Motifs    & MUTAG         & PROTEINS      & NCI1          & ABIDE         & REST-meta-MDD & Ave.          \\ \midrule
GCN          & 6.08          & 5.67          & 6.25          & 3.42          & 4.17          & 3.50          & 4.85          \\
GAT          & 6.58          & 3.25          & 5.75          & 4.58          & 3.33          & 4.67          & 4.69          \\
GIN          & {\ul 2.33}    & {\ul 3.00}    & 4.25          & {\ul 1.58}    & 7.08          & 4.25          & {\ul 3.75}    \\
SIB          & 6.17          & 7.08          & 3.75          & 6.42          & 5.83          & 4.67          & 5.65          \\
GNNExplainer & 5.25          & 4.75          & 4.42          & 5.33          & 6.00          & {\ul 3.00}    & 4.79          \\
PGExplainer  & 4.67          & 4.75          & {\ul 2.33}    & 6.67          & 3.75          & 7.33          & 4.92          \\
RC-Explainer & 3.58          & 5.92          & 7.17          & 6.58          & \textbf{2.92} & 7.08          & 5.54          \\
CI-GNN       & \textbf{1.33} & \textbf{1.58} & \textbf{2.08} & \textbf{1.42} & \textbf{2.92} & \textbf{1.50} & \textbf{1.81} \\ \bottomrule
\end{tabular}}
\end{threeparttable}
\label{tab:Performance4}
\end{table*}

\subsection{Leave-One-Site-Out Cross Validation}
To further assess the generalization ability of CI-GNN, we perform leave-one-site-out cross validations on two brain disease datasets including ABIDE and REST-meta-MDD. Both datasets encompass 17 independent sites. Specifically, each dataset is divided into the training set (16 sites out of 17 sites) and the testing set (remaining site out of 17 sites). In addition, we compare the accuracy of CI-GNN with SIB. The experimental results are presented in Table~\ref{tab:CV}.

As can be seen, CI-GNN achieves the better mean generalization accuracy of 0.71 and 0.77 than SIB and obtains noticeable improvements in most of sites on ABIDE and REST-meta-MDD, respectively. Considering that the data collected by the different sites are heterogenous, these results suggest that CI-GNN could have the potential to handle the out-of-distribution (OOD) datasets.

\begin{table}[ht!]
\centering
\caption{Leave-one-site-out cross validation on ABIDE and REST-meta-MDD. The best performance in each site is in bold.}
\begin{threeparttable}
\resizebox{\linewidth}{!}{
\begin{tabular}{@{}cccc|cccc@{}}
\toprule
\multicolumn{4}{c|}{ABIDE}                        & \multicolumn{4}{c}{REST-meta-MDD}               \\ \midrule
Site     & Sample (ASD/TD) & SIB           & CI-GNN        & Site   & Sample (MDD/HC) & SIB           & CI-GNN        \\ \midrule
CMU      & 11/13           & \textbf{0.79} & 0.75          & site1  & 73/73           & 0.59          & \textbf{0.63} \\
CALTECH  & 19/19           & 0.68          & \textbf{0.83} & site2  & 16/14           & 0.73          & \textbf{0.83} \\
KKI      & 22/33           & 0.58          & \textbf{0.82} & site3  & 18/23           & 0.83          & \textbf{0.85} \\
LEUVEN   & 29/35           & \textbf{0.64} & 0.6           & site4  & 35/37           & 0.65          & \textbf{0.76} \\
MAX\_MUN  & 24/33           & 0.63          & \textbf{0.69} & site5  & 39/48           & \textbf{0.7}  & \textbf{0.7}  \\
NYU      & 79/105          & \textbf{0.64} & \textbf{0.64} & site6  & 48/48           & 0.75          & \textbf{0.81} \\
OHSU     & 13/15           & 0.61          & \textbf{0.67} & site7  & 45/26           & 0.74          & \textbf{0.75} \\
OLIN     & 20/16           & 0.64          & \textbf{0.79} & site8  & 20/17           & 0.69          & \textbf{0.73} \\
PITT     & 30/27           & 0.61          & \textbf{0.78} & site9  & 20/16           & \textbf{0.76} & 0.72          \\
SBL      & 15/15           & 0.63          & \textbf{0.69} & site10 & 61/32           & \textbf{0.71} & 0.68          \\
SDSU     & 14/22           & 0.75          & \textbf{0.83} & site11 & 30/37           & 0.72          & \textbf{0.81} \\
STANFORD & 20/20           & 0.63          & \textbf{0.75} & site12 & 41/41           & 0.62          & \textbf{0.75} \\
TRINITY  & 24/25           & 0.63          & \textbf{0.64} & site13 & 18/31           & \textbf{0.8}  & 0.73          \\
UCLA     & 54/45           & 0.6           & \textbf{0.63} & site14 & 245/225         & 0.55         & \textbf{0.63} \\
UM       & 68/77           & \textbf{0.62} & 0.61          & site15 & 79/65           & 0.55          & \textbf{0.68} \\
USM      & 58/43           & 0.59          & \textbf{0.66} & site16 & 18/20           & 0.73          & \textbf{0.75} \\
YALE     & 28/28           & 0.61          & \textbf{0.75} & site17 & 22/23           & 0.56          & \textbf{0.64}  \\
Mean     & 31/34           & 0.64          & \textbf{0.71} & Mean   & 49/46           & 0.69         & \textbf{0.73} \\ \bottomrule
\end{tabular}}
\end{threeparttable}
\label{tab:CV}
\end{table}

\subsection {  Inner connections between different causal effect estimators including $I(\alpha;Y|\beta)$ and $I(\alpha;\hat{Y}|\beta)$}
 In this study, we use conditional mutual information (CMI) $I(\alpha;Y|\beta)$ to measure the strength of causal influence of subgraph $\alpha$ on labels $Y$. We also analyze the rationality of CMI term from a Granger causality perspective~\cite{seth2007granger}. However, previous study argue that the explanation should be derived from the classifier outputs $\hat{Y}$ rather than the labels $Y$. They leverage CMI $I(\alpha;\hat{Y}|\beta)$ to  quantify the causal influence of $\alpha$ on $\hat{Y}$. In order to further investigate the differences between the two CMI terms, we compare the impact on classification results by using $I(\alpha;Y|\beta)$ and $I(\alpha;\hat{Y}|\beta)$ on the SRPBS dataset. The experimental results are presented in Table~\ref{tab:YPerformance}. As can be seen, the performance using $(I(\alpha;Y|\beta))$ is slightly better than the performance using $(I(\alpha;\hat{Y}|\beta))$, but the difference is not significant, implying that the two methods do not have a fundamental distinction.

\begin{table*}[ht!]
\centering
\caption{ \color{blue} The classification performance and standard deviations of CI-GNN $(I(\alpha;Y|\beta))$ and CI-GNN $(I(\alpha;\hat{Y}|\beta))$ on SRPBS dataset.}
\renewcommand\arraystretch{1}
\begin{tabular}{@{}cccc@{}}
\toprule
\textbf{Method}     & \textbf{Accuracy}    & \textbf{F1}          & \textbf{MCC}         \\ \midrule
CI-GNN $(I(\alpha;Y|\beta))$                 &0.93 ± 0.03 & 0.93 ± 0.03 & 0.86 ± 0.06        \\
CI-GNN $(I(\alpha;\hat{Y}|\beta))$                & 0.91 ± 0.03          & 0.90 ± 0.04          & 0.82 ± 0.07          \\
\bottomrule
\end{tabular}
\label{tab:YPerformance}
\end{table*}



\vfill
\newpage
\section{Minimal Implementation of CI-GNN In PyTorch}

We finally provide PyTorch implementation of CI-GNN. Specifically, we show the code of the matrix-based R{\'e}nyi's $\delta$-order entropy functional as well as how to adaptively choose kernel width $\sigma$. We also show the code to evaluate the loss of causal effect estimator (i.e., Eq. (10) in the main manuscript) and to optimize $\mathcal{L}_1$ (i.e., Eq. (15) in the main manuscript). 

\begin{lstlisting}[language=Python,title={CI-GNN PyTorch}]

import torch
from scipy.spatial.distance import pdist, squareform

def reyi_entropy(k, eps=1e-8, alpha=1.01):

    k = k/(torch.trace(k)+eps)
    eigv = torch.abs(torch.linalg.eigh(k)[0])
    eig_pow = eigv**alpha
    entropy = (1/(1-alpha)) * torch.log2(torch.sum(eig_pow))
    return entropy
    
def pairwise_distances(x):

    #x should be two dimensional
    if x.dim()==1:
        x = x.unsqueeze(1)
    instances_norm = torch.sum(x**2,-1).reshape((-1,1))
    return -2*torch.mm(x,x.t()) + instances_norm + instances_norm.t()

def calculate_sigma(x):   

    x = x.cpu().detach().numpy()
    k = squareform(pdist(x, 'euclidean')) # Calculate Euclidiean distance between all samples.
    sigma = np.mean(np.mean(np.sort(k[:, :10], 1))) 

    return sigma 

def calculate_gram_mat(x):

    dist= pairwise_distances(x)
    sigma = calculate_sigma(x)**2
    return torch.exp(-dist /sigma)

def calculate_causal_loss(alpha,beta,Y):

    s_alpha = calculate_sigma_mat(alpha)
    s_beta = calculate_sigma_mat(beta)
    s_Y = calculate_sigma_mat(Y)
    #compute the causal influence of alpha on Y
    Conditional_MI = reyi_entropy(s_Y*s_beta) + reyi_entropy(s_alpha*s_beta) - reyi_entropy(s_beta) - reyi_entropy(s_alpha*s_Y*s_beta)
    #compute the mutual information between alpha and beta
    MI = reyi_entropy(s_alpha) + reyi_entropy(s_beta) - reyi_entropy(s_alpha,s_beta)
    return MI - Conditional_MI

def generative_causal(Z, Y, VAE_loss, K, lambda1):
    '''
    Args:
        Z: the unobserved latent factor obtained by GraphVAE
        Y: the true label
        VAE_loss: the objective function of GraphVAE
        K: the pre-defined feature dimensions for alpha
        lambda: the hyper-parameters
    ''' 
    
    alpha = z[:,:K]
    beta = z[:,K:]   
    causal_loss= calculate_causal_loss(alpha,beta,Y)
    #compute gradient
    loss = VAE_loss + lambda * causal_loss
    loss.backward()
    optimizer.step()
    optimizer.zero_grad()


\end{lstlisting}

\bibliography{CI-GNN.bib}
\end{document}